\pdfoutput=1
\documentclass[10pt,twocolumn,letterpaper]{article}
\usepackage[pagenumbers]{cvpr}
\usepackage{tabularx}
\usepackage{multirow}
\usepackage{balance}    %
\usepackage{placeins}   %
\definecolor{cvprblue}{rgb}{0.21,0.49,0.74}
\usepackage[pagebackref,breaklinks,colorlinks,allcolors=cvprblue]{hyperref}

\newcommand{\VO}{V}
\newcommand{\TO}{T}
\newcommand{\VT}{V{+}T}

\def\confName{CVPR}
\def\confYear{2027}

\title{TouchBench: Can Vision-Language Models Feel Invisible Physical Properties?}
\author{
Ziyao Zeng$^{1,2,\dagger}$ \quad
Ziyi Kou$^{1}$ \quad
Chuan Qin$^{1}$ \quad
Mia Huang$^{1}$ \quad
Roger Boldu$^{1}$\\
Zhiwen Fan$^{3}$ \quad
Daniel Rakita$^{2}$ \quad
Eric Whitmire$^{1}$ \quad
Li Guan$^{1}$\\[0.6em]
$^{1}$Meta Reality Labs \qquad
$^{2}$Yale University \qquad
$^{3}$Texas A\&M University\\
{\tt\small ziyao.zeng@yale.edu}
}

\begin{document}
\maketitle
{\let\thefootnote\relax\footnotetext{$^{\dagger}$Work done during an internship at Meta Reality Labs.}}

\begin{abstract}
Vision-language models (VLMs) are increasingly used for embodied perception and reasoning, yet their physical understanding is largely evaluated on questions answerable from images. Many properties critical to manipulation, such as an opaque bottle's fill level or an object's compliance, may only be revealed through touch. We introduce \textbf{TouchBench}, an egocentric vision-language-tactile benchmark that distinguishes visually answerable questions (\textbf{Tier~A}) from those involving properties hidden from the camera (\textbf{Tier~B}). It pairs head-mounted video with synchronized 137-channel whole-hand tactile sensing, comprising 900 clips from five participants across 11 tasks, with 900 absolute and 4,102 relative questions. Four state-of-the-art VLMs average $66.4\%$ accuracy on Tier~A relative questions but only $49.8\%$ on Tier~B, near chance ($50\%$). To bridge this gap, we propose \textbf{Tactile-as-Prompt} (TaP), which presents 34 interpretable tactile statistics as text, enabling VLMs to reason with touch without a learned tactile encoder. TaP raises zero-shot Tier~B relative accuracy to $62.9\%$ and reaches $73.1\%$ after leave-one-user-out fine-tuning, outperforming video alone ($67.4\%$) and a learned tactile-token encoder ($67.3\%$). The gains extend to unseen objects and users. On a separate 20-hour in-the-wild corpus, tactile text improves fine-tuned contact-question accuracy from $67.1\%$ to $75.5\%$. These results highlight the limits of vision-only physical reasoning and establish language as an effective interface for tactile perception.
\end{abstract}

\begin{figure}[t]
  \centering
  \includegraphics[width=\linewidth]{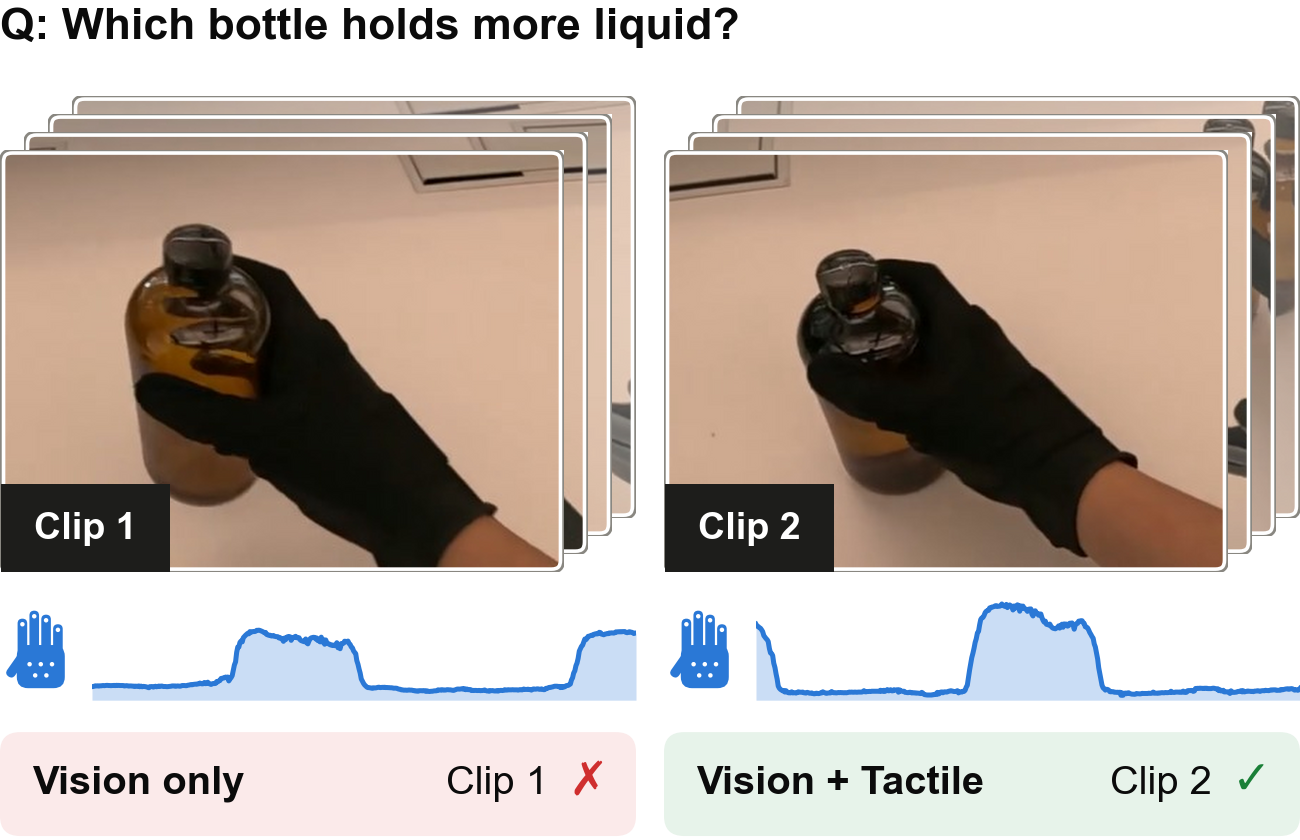}
  \caption{\textbf{Touch answers what the camera cannot.} The two amber
  bottles are hard to tell apart by eye, but only one is full. Given only the video, a
  fine-tuned VLM picks the wrong clip. Given the video and the tactile
  statistics as text, it picks the full bottle. Curves: total glove pressure.}
  \label{fig:teaser}
\end{figure}

\section{Introduction}
\label{sec:intro}

Vision-language models (VLMs) are becoming the perception and reasoning
component of embodied systems, from instruction-following manipulation
policies~\cite{palme,rt2,openvla,pi0} to embodied question
answering~\cite{embodiedqa,openeqa}. Their evaluation, however, remains
centered on the camera: benchmarks of physical and intuitive
reasoning~\cite{piqa,clevrer,intphys,physion,pacs,physbench} pose questions
that are answerable from the visual input by construction. Yet many
properties that determine how an object should be handled cannot be recovered
from an image. Two identical bottles differ in content only when lifted, two
balls of the same color differ in compliance only when squeezed and the
texture of a surface disappears once the hand covers it. Humans
resolve these through touch (\cref{fig:teaser}).

This raises two questions. Do current VLMs fail on such properties, or do
contextual cues let them answer anyway? And if they fail, in what form should
a whole-hand tactile signal enter the model?

A low score on a hard benchmark cannot settle the first question, because it
may reflect a missing percept or merely a difficult task. We therefore build
the control into the data. Every question in \textbf{TouchBench}, our
benchmark for vision-language-tactile models (VLTMs), is labeled
by visual answerability: in \textbf{Tier~A} the queried property is visible in
the video, and in \textbf{Tier~B} the capture hides it, for example behind an
opaque wall or under the hand.
When a task spans both tiers, the attribute, question and options are
shared and only the capture condition differs (e.g., a transparent
versus an opaque bottle). A model that is accurate on Tier~A but at chance on
Tier~B therefore points to a missing percept rather than to task difficulty. The
benchmark pairs egocentric video with a synchronized 137-channel tactile glove
over 900 clips from 5 participants, 11 tasks and 6 attribute dimensions, with
\emph{absolute} questions about one clip and \emph{relative} comparisons between
two clips.

Four state-of-the-art VLMs average
$66.4\%$ on Tier~A relative comparisons but only $49.8\%$ on Tier~B, where
chance is $50\%$, and the gap persists when both tiers are restricted to the
same tasks. Near-chance accuracy does not mean random answers: on Tier~B,
three of the four models choose the second clip in $71\%$ to $82\%$ of
comparisons and the fourth in $55\%$, although the correct answers are
balanced.

For the second question, the conventional approach trains an encoder that
maps tactile signals into the model's embedding
space~\cite{unitouch,tvl,octopi,sparsh}, which requires paired data that is
scarce for dexterous manipulation. We instead propose
\textbf{Tactile-as-Prompt} (TaP): each clip is summarized by 34 interpretable
statistics of force, contact geometry, vibration and wrist motion, which are
written into the text prompt. Like vision-language-action models that expose
low-dimensional state to the backbone~\cite{pi05} and work that gives sensor
readings to language models as text~\cite{penetrativeai,llmtime,hargpt}, TaP relies on the
language backbone to read physical quantities directly. It needs neither
training nor model weights, so it can apply to proprietary models.

Averaged over the four models, TaP raises zero-shot Tier~B relative accuracy
from $49.8\%$ to $62.9\%$, and tactile text alone reaches $68.7\%$. With
leave-one-user-out (LOUO) fine-tuning, video plus TaP reaches $73.1\%$, against
$67.4\%$ for video only and $67.3\%$ for learned tokens, and the
gain persists on unseen objects with an unseen user. On 20 hours of public
in-the-wild egocentric video, with $17{,}544$ evaluation questions whose
answers are measured from the tactile signal, fine-tuning with tactile text
improves accuracy from $67.1\%$ to $75.5\%$. Our contributions are:
\begin{enumerate}
  \item \textbf{TouchBench}, an egocentric video-tactile benchmark in which
    visual answerability is a controlled variable, with 900 absolute and
    4{,}102 relative questions, each labeled Tier~A or Tier~B.
  \item \textbf{A systematic evaluation of four VLMs} showing that they answer
    near chance when the queried property is hidden from the camera, with
    Tier~A as the control, and that on Tier~B their answers depend on clip
    order.
  \item \textbf{Tactile-as-Prompt}, a training-free tactile interface,
    evaluated zero-shot, after fine-tuning and on unseen objects and users,
    compared with a learned-token encoder, other training-free formats and a
    feature-only classifier, plus tactile text on a 20-hour in-the-wild
    corpus.
\end{enumerate}

\begin{figure*}[t]
  \centering
  \includegraphics[width=\linewidth]{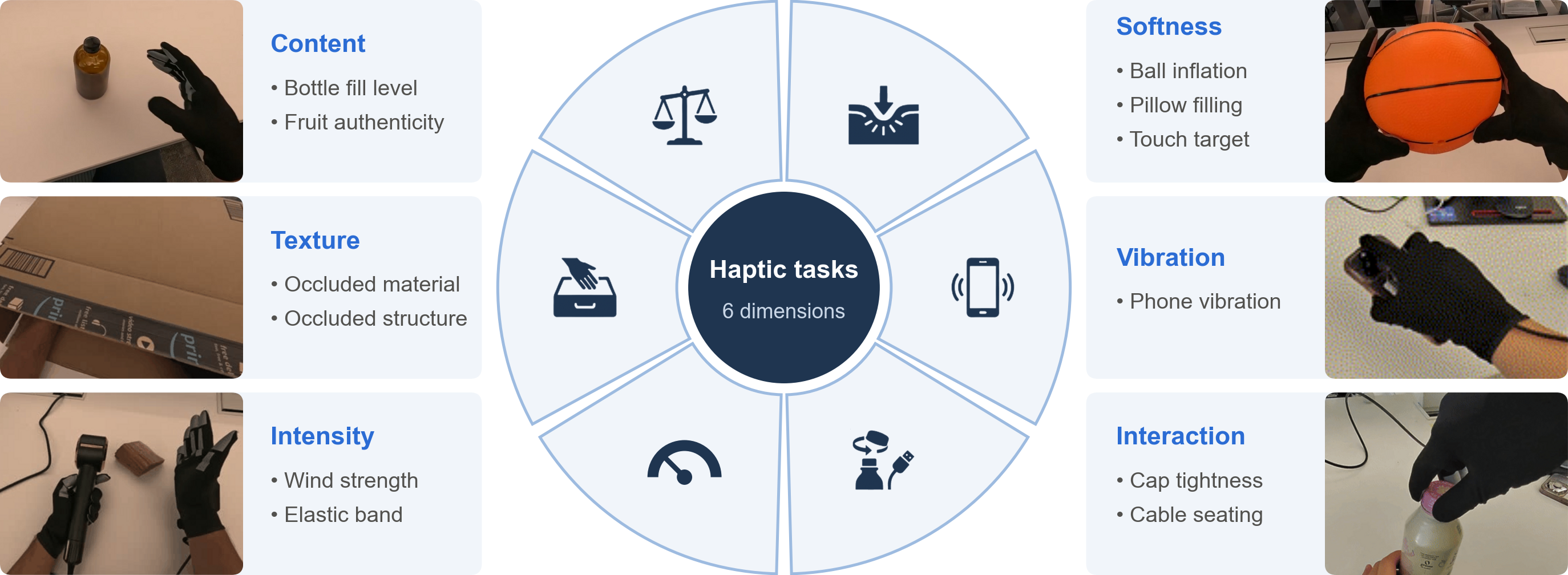}
  \caption{\textbf{The six attribute dimensions and eleven tasks of
  TouchBench.} Each task leaves the queried property visually ambiguous over
  at least part of its range: the fill level behind an opaque wall, the
  inflation of a ball that looks the same until squeezed, a surface covered by
  the hand, vibration with no visible motion, the tightness of a cap and the
  strength of wind or elastic tension. The touch-target task asks whether the
  hand presses a gloved or a bare hand, and the occluded-texture task is listed
  as its two material pairs. Clip counts per task and tier are in
  \cref{fig:stats}.}
  \label{fig:overview}
\end{figure*}

\section{Related Work}
\label{sec:related}

\paragraph{Tactile sensing and representation learning}
Vision-based tactile sensors such as GelSight~\cite{gelsight},
GelSlim~\cite{gelslim} and DIGIT~\cite{digit} image contact at high
resolution, while taxel arrays and sensorized gloves measure pressure across
the whole hand~\cite{stag}. Learned tactile representations predict grasp
success~\cite{feelingofsuccess}, fuse vision and touch for
control~\cite{makingsense}, translate between the two
modalities~\cite{crossmodal,touchandgo} and are pre-trained on large
unlabeled corpora~\cite{sparsh} or bound to vision and
language~\cite{unitouch}. We compare a learned encoder of this kind with
interpretable statistics supplied as prompt text.

\paragraph{Touch and language models}
Building on multimodal binding~\cite{imagebind}, touch-vision-language
datasets align tactile images with descriptions~\cite{tvl}, unified
embeddings connect touch to text and images~\cite{unitouch} and
tactile-language models reason about object properties from sensor
readings~\cite{octopi}. Recent systems extend tactile reasoning to ordinal
comparison and visual-tactile conflict~\cite{touchr1}, interactive
reasoning~\cite{tacreasoner} and larger multisensor
corpora~\cite{touchthinker}. All of them use learned tactile representations.
A separate line of work feeds sensor readings or time series to language models as
text~\cite{penetrativeai,hargpt,llmtime}, composes pretrained models through
language~\cite{socratic}, summarizes multisensory robot observations as text
for failure reasoning~\cite{reflect} or aligns motion-sensor embeddings with
video and language~\cite{imu2clip}. Relatedly, vision-language-action models expose
proprioceptive state to the backbone~\cite{pi05}. We compare a learned-token
interface and a statistics-as-text interface on the same supervised data. Text
also enables tactile evaluation of proprietary models.

\paragraph{Physical and embodied reasoning benchmarks}
Benchmarks probe physical commonsense in language~\cite{piqa}, intuitive
physics from video~\cite{clevrer,intphys,physion}, audiovisual
commonsense~\cite{pacs}, embodied question
answering~\cite{embodiedqa,openeqa} and physical understanding in
VLMs~\cite{physbench}. Their questions are answerable from the perceptual
input, so a low score cannot separate a missing percept from failed
reasoning. TouchBench supplies this control by partitioning questions by
visual answerability. Touch-R1~\cite{touchr1} uses a complementary control,
crediting tactile reasoning only when real tactile input outperforms
counterfactual inputs.

\paragraph{Egocentric video and hand-object interaction}
Egocentric corpora such as Ego4D~\cite{ego4d}, Ego-Exo4D~\cite{egoexo4d} and
EPIC-KITCHENS~\cite{epic} capture manipulation at scale, and hand-object
datasets add contact detection~\cite{hoi100doh}, grasp
capture~\cite{dexycb} and per-vertex contact~\cite{contactpose,grab}. A few
recordings pair egocentric video with tactile
gloves~\cite{actionsense,egotouch}, and recent work learns to predict
hand pressure or contact from egocentric video
alone~\cite{egotactile,egotac,touchsight}. None of these aligns dense
per-finger tactile data with question-answer labels sorted by visual
answerability. Moreover, predicting touch from video cannot recover properties that
leave no visual trace, which Tier~B targets.

\section{TouchBench}
\label{sec:bench}

\subsection{Data collection}
\label{sec:capture}
Five participants manipulated
everyday objects while we recorded egocentric
video from a head-mounted camera and tactile data from a sensorized glove on
the right hand. The glove records \textbf{137 channels} at about 98\,Hz, comprising 132 pressure taxels
(60 on the fingertips, 72 on the palm) and 5 per-finger flexion channels. A wrist-mounted inertial measurement unit also reports
orientation as a quaternion. Video and glove streams are synchronized in software
via a timestamp code displayed in the video. The participants, P1 to P5,
contributed 75 minutes of manipulation and \textbf{900 synchronized clips},
each subsampled to 32 frames for model input. 

\subsection{Tasks and questions}
\label{sec:questions}
The eleven tasks are organized into six attribute dimensions
(\cref{fig:overview}). Scoring the occluded-texture task as two material
pairs gives twelve scoring dimensions. Each clip supports two question
formats. An \emph{absolute} question queries a single clip (e.g., ``how tightly is the cap screwed on?'') with two options (four for the elastic-band count). A \emph{relative} question compares two clips from the same dimension (e.g., ``which pillow is softer?'') with two options, yielding a $50\%$ chance level. The benchmark contains \textbf{900 absolute} and
\textbf{4{,}102 relative} questions. Pairs may span visibility conditions
within a dimension. These 922 mixed-tier pairs are excluded from
tier-specific results (\cref{tab:micro}).

\begin{figure*}[t]
  \centering
  \includegraphics[width=\linewidth]{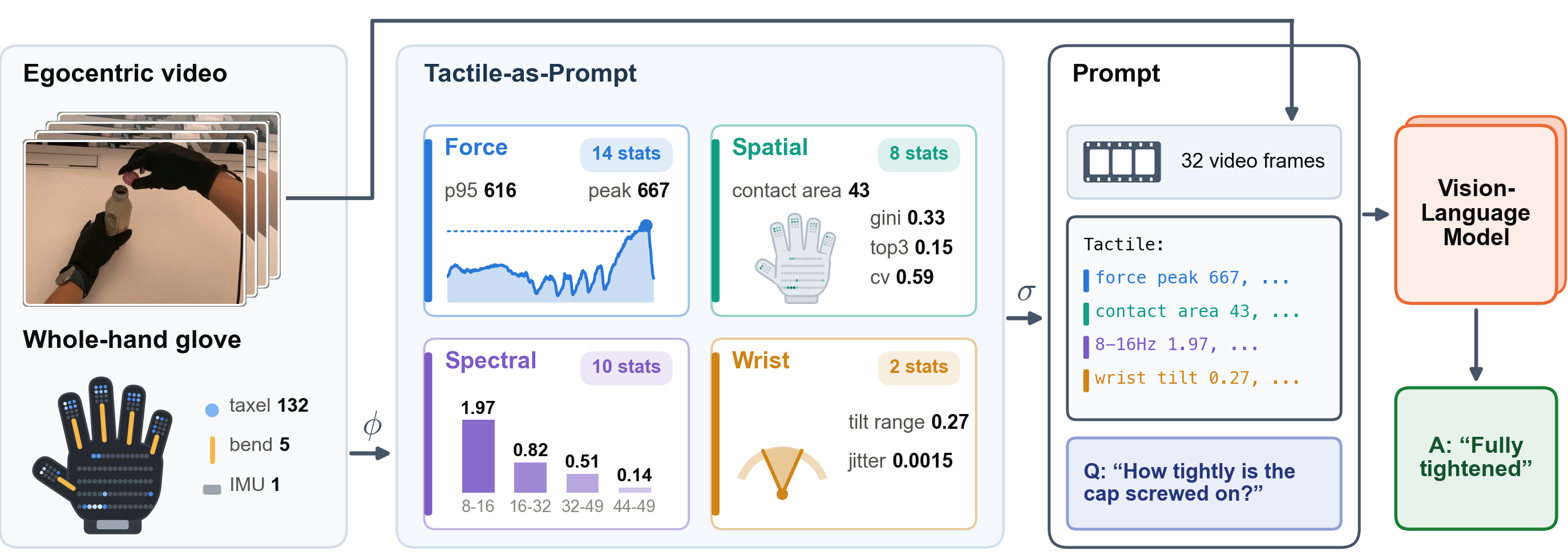}
  \caption{\textbf{The Tactile-as-Prompt interface} can turn the input whole-hand touch into a prompt for VLM inference. The glove stream is summarized by 34 interpretable
  statistics in four groups ($\phi$) and serialized as text into the prompt
  ($\sigma$). For illustration, each card and prompt shows a few of its group's
  statistics with their values for this clip. The video frames, this line, and the question form a prompt for VLM inference.
  No training is needed so the same
  prompt can be sent to an open-weight or a proprietary model.}
  \label{fig:pipeline}
\end{figure*}

\subsection{Visual answerability: Tier A and Tier B}
\label{sec:tiers}
Every question is labeled by whether its target property is visible in the
video. In \textbf{Tier~A}, the property is visible, such as the texture of an unoccluded
surface or the fill level of a transparent bottle. In \textbf{Tier~B}, the
capture condition hides it, such as the same texture under the hand or the
fill level of an opaque bottle. Tiers therefore follow from the capture
design and were assigned once per capture subcategory. A capture subcategory is labeled Tier~A only
when its target property can be determined from the pixels without relying on
object identity or another statistical proxy. We audited the subcategories
with suspicious zero-shot scores by answering their items from the frames,
question and options alone. In the Tier~A structure pair, where
video-only models score below chance, the 8 items that could be judged agree
with the gold answer, so neither labels nor wording are inverted. The
semi-transparent amber bottle shows no fill difference in any inspected frame
and was moved to Tier~B. For the elastic bands, the thickness of the
stretched bundle is a partial visual cue (13 of 20 pairs judged correctly
with low confidence), but the count itself is not visible, so the task stays
in Tier~B. A visible control pair was judged correctly in all 10 items, and no gold label was changed (\cref{app:audit}). Five scoring dimensions (bottle, glove, phone and the two
texture pairs) contain both tiers and seven contain only Tier~B, so aggregate
tier comparisons also reflect task composition. \cref{sec:vlmfail} therefore
compares the tiers within the shared dimensions as well.

Tier~A is a control rather than a target: a model that fails on Tier~A fails
at the task itself. Conversely, a Tier~B label does not prevent
a model from learning a visual proxy, such as the thickness of a stretched
band bundle, which \cref{sec:vlmfail} suggests happens after fine-tuning.

\subsection{Evaluation protocol}
\label{sec:protocol}
We evaluate three input conditions, video only (\VO), tactile only (\TO) and
both (\VT), in four settings. \emph{Zero-shot} evaluation uses all items.
Tier-specific relative results use the 3{,}180 same-tier pairs (543 Tier~A
and 2{,}637 Tier~B), and absolute items comprise 218 Tier~A and 682 Tier~B
examples. \emph{Leave-one-user-out} (LOUO) fine-tuning trains on four
participants and evaluates on the fifth. Training is capped at 48 pairs per
scoring dimension, and no evaluation clip or source segment appears in training.
Pooling the five folds yields the same items as the zero-shot rows. The
\emph{unseen-object} setting tests on 225 additional clips that one
participant recorded with new object instances, while training only on the other
participants. The \emph{web-scale} setting uses questions constructed
without human annotation (\cref{sec:scale}). Unless noted otherwise, accuracy
is the unweighted mean over capture subcategories (6 in Tier~A, 16 in
Tier~B), with each relative pair attributed to its first clip, so that
subcategories with many pairs do not dominate the average. Item-weighted
results for the main comparisons, given in \cref{tab:micro}, preserve the
pattern. Chance is $50.0\%$ for relative
questions. For absolute questions, chance is $50.0\%$ on Tier~A and $48.4\%$ on Tier~B, because the four-option band count is a Tier~B task. Where an
answer-wording prior exists (\cref{sec:scale}), we report it alongside
chance.

\section{Tactile-as-Prompt}
\label{sec:method}

\subsection{Problem formulation}
A clip is a pair $(\mathcal{V},\mathcal{T})$ of 32 RGB frames
$\mathcal{V}=\{I_1,\dots,I_{32}\}$ and a glove stream
$\mathcal{T}\in\mathbb{R}^{L\times 137}$ with a wrist quaternion track
$\mathcal{Q}\in\mathbb{R}^{L\times 4}$. Given a question $q$ with options
$\mathcal{O}$, the model predicts an answer $\hat{a}\in\mathcal{O}$. A tactile interface
maps $(\mathcal{T},\mathcal{Q})$ into a form the language backbone can
consume. A learned-token interface trains an encoder
$E_\theta:(\mathcal{T},\mathcal{Q})\rightarrow\mathbb{R}^{k\times d}$ into the
$d$-dimensional embedding space and must learn $\theta$ from paired
supervision. \textbf{Tactile-as-Prompt} instead composes a fixed feature map
with a fixed serializer,
\begin{equation}
  \sigma\big(\phi(\mathcal{T},\mathcal{Q})\big),\qquad
  \phi:\mathbb{R}^{L\times 137}\!\times\!\mathbb{R}^{L\times 4}
  \rightarrow\mathbb{R}^{34},
\end{equation}
where $\sigma$ renders the 34 values as one line of text. The prompt is
$p=[\text{instruction};\,\mathcal{V};\,\sigma(\phi(\mathcal{T},\mathcal{Q}));\,
q;\,\mathcal{O}]$. For a relative question, the video and tactile blocks of
both clips precede the question. \VO{} omits the tactile block and \TO{} omits the
frames. The question and options remain identical, and the system
line specifies the supplied modalities.

\subsection{Tactile descriptors}
\label{sec:features}
The 34 statistics are computed per clip from the baseline-subtracted glove
channels and the wrist quaternion, in four groups (\cref{fig:pipeline}).
\emph{Force} (14) covers the peak, 95th percentile, mean and spread of total
pressure, the peak of each fingertip pad, the fingertip and palm totals and
their ratio, and flexion. \emph{Spatial} (8) covers contact area,
concentration and the temporal profile of the grasp. \emph{Spectral} (10)
covers Welch band powers between 8 and 49\,Hz, the spectral centroid and a
jitter term, and \emph{Wrist} (2) covers wrist stability. Whole-hand totals
sum all channels, including the bend channels, so they are signal summaries
in sensor units rather than calibrated forces. Values are written at fixed
precision and in a fixed order with units and plain names, for example
``contact area 41 taxels, concentration gini 0.62'', so the same string can
be sent to any model. The statistics were designed on a pilot recording of
one participant, whose clips later form the P5 fold. \Cref{app:features} gives full
definitions.

\subsection{A learned-token alternative}
\label{sec:tokens}
As a conventional alternative, we also train a learned tactile encoder using the same supervised data, backbone (Qwen3.5-9B with LoRA) and schedule. A 1.79M-parameter temporal convolution
encoder maps the glove stream to $k{=}32$ tokens, projected through a
learned gate into the language embedding space. The encoder is pre-trained via frame-level contrastive alignment with frozen visual features on 50 minutes of our free-form manipulation data and the 20-hour public corpus of \cref{sec:scale}, then fine-tuned with the backbone. Its configuration
was selected among fusion, gating, normalization and pre-training variants, so
we treat it as a reference baseline rather than a tuned optimum (\cref{app:training}).

\section{Experiments}
\label{sec:exp}

\begin{table*}[t]\centering\small
\setlength{\tabcolsep}{3.2pt}
\begin{tabular}{@{}l ccc ccc c ccc ccc@{}}
\toprule
& \multicolumn{6}{c}{\textbf{Relative} (two clips, chance $50.0$)}
& & \multicolumn{6}{c}{\textbf{Absolute} (one clip, chance $50.0$ / $48.4$)} \\
\cmidrule(lr){2-7}\cmidrule(lr){9-14}
& \multicolumn{3}{c}{Tier~A \ ($n{=}543$)}
& \multicolumn{3}{c}{Tier~B \ ($n{=}2{,}637$)}
& & \multicolumn{3}{c}{Tier~A \ ($n{=}218$)}
& \multicolumn{3}{c}{Tier~B \ ($n{=}682$)} \\
\cmidrule(lr){2-4}\cmidrule(lr){5-7}\cmidrule(lr){9-11}\cmidrule(lr){12-14}
Model & \VO & \TO & \VT & \VO & \TO & \VT & & \VO & \TO & \VT & \VO & \TO & \VT \\
\midrule
\multicolumn{14}{@{}l}{\emph{Zero-shot}} \\
InternVL3-8B     & 70.6 & 55.9 & \textbf{77.0} & 50.0 & \textbf{64.2} & 53.6 & & 58.3 & 50.0 & \textbf{60.2} & 50.6 & 48.6 & \textbf{52.3} \\
Qwen3.5-9B       & \textbf{74.3} & 57.3 & 73.4 & 49.8 & \textbf{69.7} & 62.3 & & \textbf{69.6} & 48.0 & 63.4 & \textbf{50.1} & 48.7 & \textbf{50.1} \\
Claude Sonnet    & 60.6 & 58.3 & \textbf{63.5} & 50.8 & \textbf{69.7} & 66.9 & & \textbf{64.6} & 51.8 & 64.0 & 48.9 & 48.3 & \textbf{50.3} \\
Claude Opus      & 60.1 & 58.0 & \textbf{64.9} & 48.7 & \textbf{71.2} & 68.7 & & \textbf{71.1} & 48.1 & 66.8 & 51.4 & 48.2 & \textbf{51.9} \\
\addlinespace[2pt]
\emph{mean}      & \emph{66.4} & \emph{57.4} & \emph{69.7} & \emph{49.8} & \emph{68.7} & \emph{62.9} & & \emph{65.9} & \emph{49.5} & \emph{63.6} & \emph{50.2} & \emph{48.4} & \emph{51.1} \\
\midrule
\multicolumn{14}{@{}l}{\emph{Fine-tuned, leave-one-user-out (Qwen3.5-9B)}} \\
video only                      & \textbf{99.5} & --   & --   & 67.4 & --   & --   & & 97.1 & --   & --   & 63.3 & --   & -- \\
\quad + learned tokens          & --   & 48.7 & 98.9 & --   & 53.5 & 67.3 & & --   & 55.4 & 97.9 & --   & 50.0 & 59.8 \\
\quad + TaP (ours)              & --   & 56.7 & 97.6 & --   & 69.4 & \textbf{73.1} & & --   & 49.7 & \textbf{99.2} & --   & 49.3 & \textbf{66.5} \\
\bottomrule
\end{tabular}
\caption{\textbf{Accuracy (\%) on TouchBench by question format, tier and
input condition.} Values are macro-averages over capture subcategories, and $n$ is the number
of items. \VO: video only. \TO: tactile only. \VT: both. With video alone, the
four models average $66.4$ (relative) and $65.9$ (absolute) on Tier~A, but
every model stays within $3.0$ points of chance on Tier~B in both formats.
Tactile evidence, serialized as text, recovers Tier~B. The fine-tuned
block shares one video-only model across both interfaces. Bold marks the best
input condition per row and tier for the zero-shot models and the best
fine-tuned arm per tier.}
\label{tab:main}
\end{table*}

\subsection{Setup}
We evaluate InternVL3-8B~\cite{internvl3} and
Qwen3.5-9B~\cite{qwenvl}, and two proprietary models, Claude Sonnet and
Claude Opus (\texttt{claude-sonnet-5} and \texttt{claude-opus-5}). The InternVL3-8B and Qwen3.5-9B models receive 32 frames per clip and are scored by the option letter with
the largest next-token logit, with thinking disabled. The Claude models,
queried on 14 October 2026, receive the frames of each clip as two contact sheets in the user message,
with all tools disabled, and are prompted for a single letter at default
sampling settings. All of their responses are valid. Fine-tuning freezes
Qwen3.5-9B and trains LoRA adapters ($r{=}32$, $\alpha{=}64$, dropout $0.05$)
on the attention and MLP projections with AdamW (peak learning rate
$2\times10^{-4}$, linear schedule, $3\%$ warm-up, bfloat16, effective batch
size 16), for five epochs on relative and ten on absolute questions, on one
H100 per run. Benchmark entries are one run per arm, and
unseen-object entries average three seeds (two for tactile only). \Cref{app:training,app:prompts} give details
and prompts.

\begin{figure}[t]
  \centering
  \includegraphics[width=\linewidth]{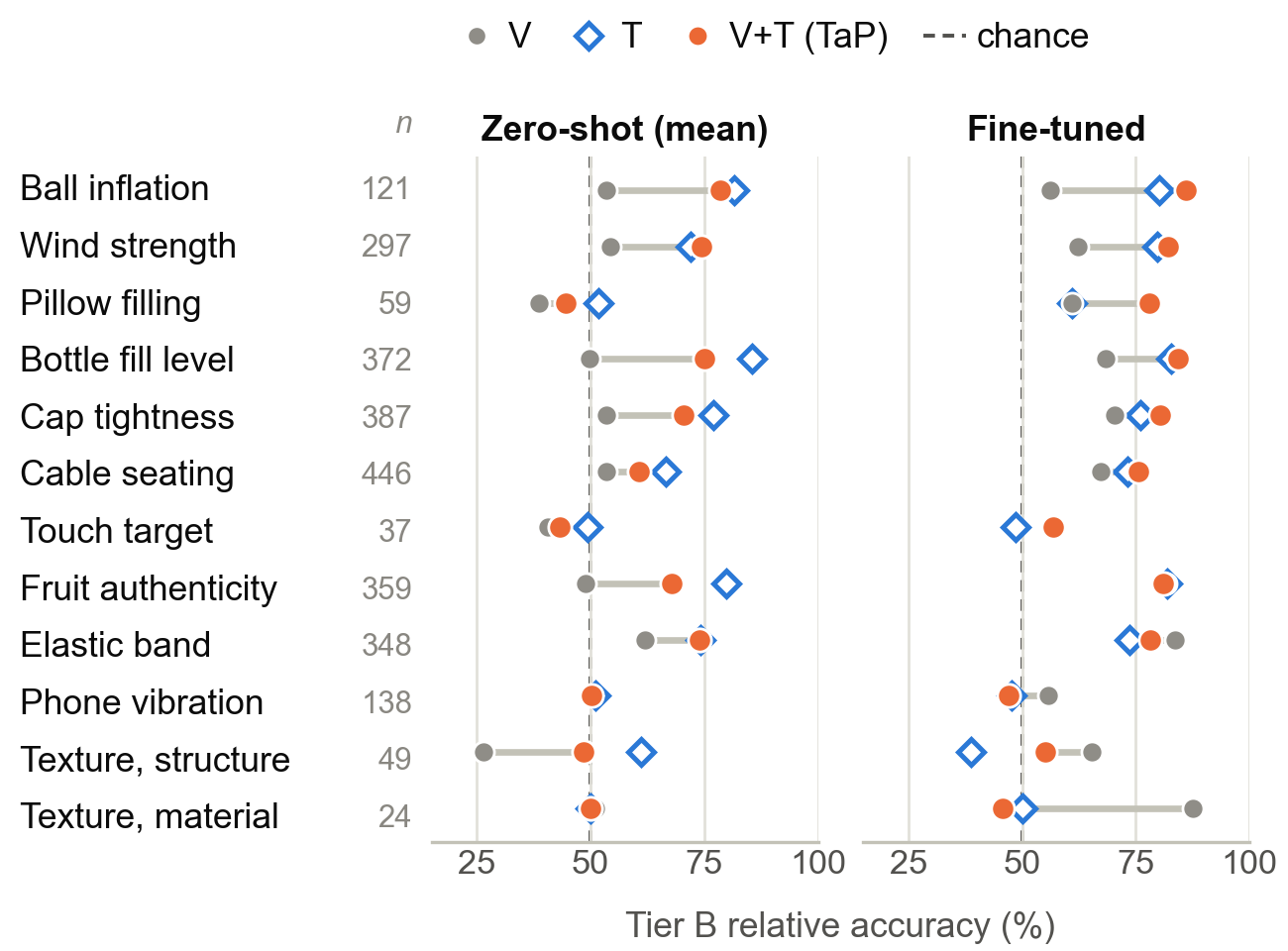}
  \caption{\textbf{Where touch helps.} Tier~B relative accuracy (\%) per
  scoring dimension, sorted by the fine-tuned gain of \VT{} over \VO. Left:
  mean of the four zero-shot models. Right: fine-tuned Qwen3.5-9B (LOUO).
  $n$ is the number of Tier~B pairs, and accuracy is item-weighted within each
  dimension.}
  \label{fig:dims}
\end{figure}

\begin{table}[t]\centering\small
\setlength{\tabcolsep}{4pt}
\begin{tabular}{@{}lcccc@{}}
\toprule
& \multicolumn{2}{c}{Shared dimensions} & \multicolumn{2}{c}{Tier~B, both orders} \\
\cmidrule(lr){2-3}\cmidrule(lr){4-5}
\VO{} model & Tier~A & Tier~B & same clip & accuracy \\
\midrule
InternVL3-8B  & 70.6 & 47.6 & 38.5 & 50.7 \\
Qwen3.5-9B    & 74.3 & 43.4 & 47.4 & 50.6 \\
Claude Sonnet & 60.6 & 46.0 & 32.2 & 51.7 \\
Claude Opus   & 60.1 & 46.4 & 48.8 & 48.0 \\
\emph{mean}   & \emph{66.4} & \emph{45.9} & \emph{41.7} & \emph{50.3} \\
\midrule
Fine-tuned    & 99.5 & 62.9 & 75.1 & 66.9 \\
\bottomrule
\end{tabular}
\caption{\textbf{Video-only controls} (relative, macro \%). \emph{Shared
dimensions}: the five scoring dimensions with both tiers. \emph{Both orders}:
each Tier~B pair is evaluated in both clip orders; \emph{same clip} is the
share of pairs answered with the same clip both times (position-only
answering gives $0$).}
\label{tab:vcontrols}
\end{table}

\subsection{Do VLMs fail on visually ambiguous properties?}
\label{sec:vlmfail}
With video alone, the four models average $66.4\%$ on Tier~A relative
comparisons and $49.8\%$ on Tier~B, where chance is $50\%$ (\cref{tab:main}).
Absolute questions show the same pattern ($65.9\%$ against $50.2\%$, with
chance at $50.0\%$ and $48.4\%$). Because the tiers differ in task composition,
we also restrict evaluation to the five scoring dimensions shared by both tiers,
and the gap persists (\cref{tab:vcontrols}): $66.4\%$ on Tier~A against $45.9\%$ on Tier~B.

\paragraph{Answer priors on Tier~B}
Near-chance accuracy does not mean random answers. Pair orientation was
randomized, so the Tier~B labels are balanced (A/B $=50.7/49.3\%$), yet the
video-only arms choose B, the second clip, in $71.4\%$ (Qwen), $54.6\%$
(InternVL), $81.6\%$ (Sonnet) and $71.9\%$ (Opus) of cases. Evaluating each
pair in both clip orders (\cref{tab:vcontrols}) confirms item-level order
dependence: the models choose the same clip in both orders for only $32$--$49\%$ of Tier~B
pairs, and their order-averaged accuracy is $48.0$--$51.7\%$. After
fine-tuning, the same clip is chosen in both orders for $75.1\%$ (video) and
$84.6\%$ (TaP) of Tier~B pairs.

\paragraph{Visual proxies after fine-tuning}
The fine-tuned video-only model reaches $67.4\%$ on Tier~B, well above
chance, with the largest gains over zero-shot Qwen3.5-9B on the two texture pairs
($41.7\%$ to $87.5\%$ and $28.6\%$ to $65.3\%$) and fruit ($50.1\%$ to
$81.6\%$). We hypothesize that fine-tuning exploits visual correlates that a
human would not consider direct evidence of the attribute, such as the thickness of
the stretched band bundle (\cref{app:audit}). This is why the fine-tuned video-only model, not the zero-shot
one, is the baseline for TaP, and why the unseen-object split matters.

\subsection{Does Tactile-as-Prompt close the gap?}
\label{sec:results}
Adding the tactile block raises mean zero-shot Tier~B relative accuracy from
$49.8\%$ to $62.9\%$, improving every model. After LOUO fine-tuning, TaP
reaches $73.1\%$ against $67.4\%$ for video alone (95\% bootstrap intervals
of the differences: $+11.4$ to $+14.7$ points zero-shot and $+3.0$ to $+8.4$
after fine-tuning; McNemar $p<10^{-11}$ for every model). Both analyses treat items
as independent, whereas pairs share clips and participants, making the intervals
and $p$-values optimistic. On Tier~A, where the
camera already supplies the answer, TaP does not help ($97.6\%$ against
$99.5\%$). On the 922 mixed-tier pairs, fine-tuned TaP reaches $92.7\%$ against $87.9\%$ (item-weighted). The fine-tuned model reads the tactile content (\cref{tab:controls}): its Tier~B
accuracy falls from $72.9\%$ to $59.8\%$ when each clip's tactile text is
replaced by that of another clip, and to $63.9\%$ when the text is withheld.

\begin{table}[t]\centering\small
\setlength{\tabcolsep}{4pt}
\begin{tabular}{@{}llccc@{}}
\toprule
Model & Input & real & shuffled & withheld \\
\midrule
\multirow{2}{*}{Fine-tuned Qwen} & \TO{} (TaP) & 71.7 & 54.5 & 48.4 \\
 & \VT{} (TaP) & 72.9 & 59.8 & 63.9 \\
\addlinespace[2pt]
\multirow{2}{*}{Zero-shot Qwen} & \TO & 69.8 & 50.1 & 50.8 \\
 & \VT & 61.8 & 50.5 & 49.9 \\
\addlinespace[2pt]
\multirow{2}{*}{Zero-shot InternVL} & \TO & 64.2 & 49.1 & 49.4 \\
 & \VT & 53.6 & 51.3 & 50.0 \\
\bottomrule
\end{tabular}
\caption{\textbf{Tactile controls at inference} (Tier~B relative, macro
accuracy \%). \emph{Shuffled}: each clip's tactile line is replaced by that of
another evaluation clip. \emph{Withheld}: the line is removed. All conditions
were re-evaluated in one software environment, so \emph{real} differs from
\cref{tab:main} by up to $2.3$ points.}
\label{tab:controls}
\end{table}

\paragraph{Tactile-only and multimodal prompting}
On zero-shot Tier~B, tactile text alone beats video plus tactile text for all
four models ($68.7\%$ against $62.9\%$ on average), whereas on Tier~A it is
the weakest input ($57.4\%$, against $66.4\%$ for video and $69.7\%$ for
both). Since every model
receives the frames, unread images do not explain the lower \VT{} accuracy on
Tier~B, but the zero-shot results alone do not identify its cause. After
fine-tuning, video plus tactile exceeds tactile alone ($73.1\%$ against
$69.4\%$).

\paragraph{Where touch helps, and where it does not}
\cref{fig:dims} breaks Tier~B down by scoring dimension. Touch helps most
when the property manifests in how the object resists or pushes back against the
hand: after fine-tuning, TaP gains $29.8$ points on ball inflation, $19.9$ on
wind strength, $17.0$ on pillow filling and $16.1$ on bottle fill level. It
does not help in three groups of dimensions. Phone vibration stays at chance
($47.1\%$), consistent with the glove's 49\,Hz Nyquist limit. On fruit
authenticity and the elastic band, fine-tuned video alone is already strong
($81.6\%$ and $83.6\%$), consistent with the visual proxies of
\cref{sec:vlmfail}, and touch does not improve accuracy ($81.1\%$ and $78.2\%$). On the two
occluded texture pairs and the touch target, the statistics carry little
information (tactile only $50.0\%$, $38.8\%$ and $48.6\%$ after fine-tuning),
and on the occluded material pair, adding touch lowers accuracy from $87.5\%$
to $45.8\%$. These dimensions are small, with 24, 49 and 37 Tier~B pairs. Four of the
five dimensions present in both tiers fall into these groups, so on the
Tier~B subset of the shared dimensions, fine-tuned TaP and video only perform comparably
($62.2\%$ and $62.9\%$ macro, see \cref{tab:shared-dims}). The fine-tuned gain comes from the
dimensions unique to Tier~B.

\subsection{Relative and absolute questions behave differently}
\label{sec:relabs}
Question format matters as much as modality. On absolute Tier~B questions,
TaP moves the fine-tuned model from $63.3\%$ to $66.5\%$, a non-significant
difference (McNemar $p=0.47$), and the zero-shot average only from
$50.2\%$ to $51.1\%$. The model uses touch far more effectively to
\emph{compare} two clips than to \emph{calibrate} a judgment: an unnormalized
peak-force reading identifies the firmer of two balls but not whether a ball
is firm. We report this negative result rather than restricting the
benchmark to the format on which the method succeeds.

\begin{figure}[t]
  \centering
  \includegraphics[width=\linewidth]{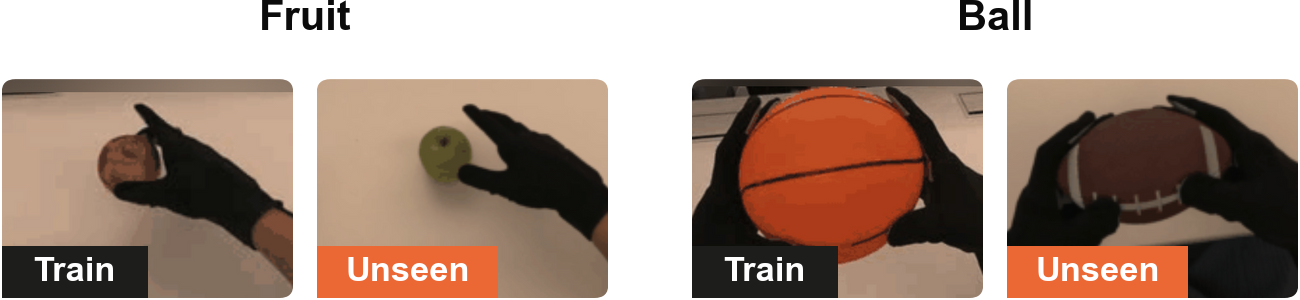}
  \caption{\textbf{Unseen objects.} Training and evaluation use different object
  instances, and the evaluation participant is held out.}
  \label{fig:unseen}
\end{figure}

\subsection{Generalization to unseen objects and users}
\label{sec:unseen}
\cref{tab:newobj} evaluates on 225 additional clips that one participant
recorded with object instances never seen in training, such as a different
apple or a football instead of a basketball (\cref{fig:unseen}). The model is
trained on the benchmark clips of the other participants. Relative comparison
improves with tactile input in all three seeds, by $4.2$ points on Tier~B and
$4.8$ points overall (item-weighted), while absolute questions improve in
none. Under the dimension macro-average the Tier~B gain shrinks to $1.7$
points ($59.1\%$ to $60.8\%$), and tactile-only text is highest ($63.3\%$;
\cref{app:unseen}).

\begin{table}[t]\centering\small
\setlength{\tabcolsep}{4pt}
\begin{tabular}{@{}llcccc@{}}
\toprule
Form & Tier & \VO & \TO & \VT & \VT$-$\VO \\
\midrule
\multirow{3}{*}{Relative}
 & A       & 97.2 & 47.9 & \textbf{97.9} & $+0.7$ \\
 & B       & 64.1 & 65.5 & \textbf{68.4} & $+4.2$ \\
 & overall & 70.1 & 62.7 & \textbf{74.9} & $\mathbf{+4.8}$ \\
\addlinespace[2pt]
\multirow{3}{*}{Absolute}
 & A       & \textbf{73.2} & 38.0 & 66.7 & $-6.5$ \\
 & B       & \textbf{55.1} & 45.5 & 53.1 & $-2.0$ \\
 & overall & \textbf{58.8} & 44.0 & 55.9 & $-3.0$ \\
\bottomrule
\end{tabular}
\caption{\textbf{Unseen objects with an unseen user} (fine-tuned
Qwen3.5-9B; item-weighted accuracy over three seeds for \VO{} and \VT{} and
two for \TO). Relative comparison improves with tactile input in 3 of 3
seeds, absolute questions in 0 of 3. Chance is $50.0$, except $47.1$ for
absolute Tier~B. $n$ is 95 / 653 relative and 46 / 179 absolute items in
Tier~A / Tier~B. The overall relative row adds 158 mixed-tier pairs.
Differences are computed before rounding.}
\label{tab:newobj}
\end{table}

\subsection{Which interface? Text, tokens, or neither}
\label{sec:interface}
\cref{tab:interface} compares the two tactile interfaces. The learned-token
encoder is weaker in both settings where it was trained. On
TouchBench it reaches $67.3\%$ against $73.1\%$ for TaP (McNemar
$p<10^{-11}$), despite its trained encoder, gate and tactile
pre-training. On the web-scale corpus, it reaches $67.8\%$ against $75.5\%$,
only $0.7$ points above video alone. There, however, it was trained from
scratch, and a control shows that it ignores its tactile input ($67.4\%$ with
real, shuffled or withheld input), likely because its tokens have about 67
times the mean norm of the model's input embeddings. These results favor statistics as text
for the encoders we tested, not over every learned encoder.

Two training-free alternatives, one rendered tactile image per clip and raw
5\,Hz pressure values as text, fall below TaP for the open models (Tier~B,
touch only, Qwen3.5-9B: $69.7\%$ against $56.4\%$ and $65.7\%$; \cref{tab:t16}). For the
Claude models, the rendered image matches TaP with touch alone but not with
video, so the advantage of statistics is model-dependent. A
gradient-boosted tree on the 34 features reaches $70.3\%$ on the
unseen-object relative split, above tactile-only text ($62.7\%$) but below
video plus tactile text ($74.9\%$).

\begin{table}[t]\centering\small
\setlength{\tabcolsep}{3pt}
\begin{tabular}{@{}lccccc@{}}
\toprule
& \multicolumn{2}{c}{Tactile only} & \multicolumn{2}{c}{Video + tactile} & \\
\cmidrule(lr){2-3}\cmidrule(lr){4-5}
Setting & tokens & text & tokens & text & \VO \\
\midrule
TouchBench, Tier~B rel. & 53.5 & 69.4 & 67.3 & \textbf{73.1} & 67.4 \\
Unseen objects, rel.\ all & --   & 62.7 & --   & \textbf{74.9} & 70.1 \\
Web-scale contact QA & 62.6 & 66.1 & 67.8 & \textbf{75.5} & 67.1 \\
\bottomrule
\end{tabular}
\caption{\textbf{Interface comparison with shared supervised data and
backbone} (accuracy \%; all rows fine-tuned). The TouchBench row is
macro-averaged as in \cref{tab:main}, and the unseen-object row is item-weighted as
in \cref{tab:newobj}. \emph{Text} is TaP on TouchBench and a per-frame
tactile table on the web-scale corpus (\cref{sec:scale}), where the token
encoder is trained from scratch and ignores its tactile input.}
\label{tab:interface}
\end{table}

\begin{table}[t]\centering\small
\setlength{\tabcolsep}{5pt}
\begin{tabular}{@{}lccc@{}}
\toprule
Descriptors & Tier~A & Tier~B & Shuffle drop (B) \\
\midrule
All 34            & 97.6  & 72.9          & 13.0 \\
$-$ Force (14)    & 99.9  & \textbf{68.4} & \textbf{1.5} \\
$-$ Spatial (8)   & 98.5  & 72.9          & 12.7 \\
$-$ Spectral (10) & 100.0 & 73.3          & 11.9 \\
$-$ Wrist (2)     & 99.6  & 71.8          & 11.4 \\
\bottomrule
\end{tabular}
\caption{\textbf{Descriptor-group ablation} (fine-tuned \VT{} TaP, LOUO,
relative macro accuracy \%). Each row retrains without one group.
\emph{Shuffle drop}: Tier~B accuracy lost, in points, when tactile lines are
shuffled across clips. The \emph{All 34} row re-evaluates the main model in the
software environment of the ablation runs ($72.9$ against $73.1$ in
\cref{tab:main}).}
\label{tab:ablation}
\end{table}

\paragraph{Descriptor-group ablation}
\label{sec:ablation}
Retraining the LOUO \VT{} TaP model without one descriptor group at a time
(\cref{tab:ablation}) shows that force carries most of the signal: without
it, Tier~B accuracy falls from $72.9\%$ to $68.4\%$ and the shuffle drop from
$13.0$ to $1.5$ points, whereas removing any other group leaves both nearly
unchanged. In free-form captioning (\cref{app:caption}), the small zero-shot gain
from touch vanishes after fine-tuning.

\subsection{Scaling to annotation-free web data}
\label{sec:scale}
To test tactile text at scale and with a different glove, we construct contact questions
automatically from 20 hours of in-the-wild manipulation in EgoTouch, a public
egocentric tactile corpus released with TouchAnything~\cite{egotouch}. A
Claude model segments each recording into captioned events, and each
question asks which of two events within a five-second clip has the larger
mean or peak force, mean or peak contact area, or duration, with ``About the
same'' for near ties. The answers are thus measured from the tactile signal
rather than generated by a model. Splitting by recording yields $104{,}210$
training and $17{,}544$ evaluation questions. Because the public glove
differs from ours, the tactile text here is a 32-frame table of left- and
right-hand grip force, contact area and contact flag. Since the answers are
computed from these same signals, this setting tests whether a model can find
and compare the two events in the tactile text, not whether it can infer a
hidden property. 

Fine-tuning with this tactile text raises accuracy from $67.1\%$ to $75.5\%$, $8.4$ points
above fine-tuning on video alone (\cref{tab:interface}), and the model uses
the tactile text: on 4{,}000 evaluation items, accuracy falls from $75.6\%$
to $49.7\%$ when it is shuffled across clips and to $58.9\%$ when it is
withheld.

\begin{figure}[t]
  \centering
  \includegraphics[width=\linewidth]{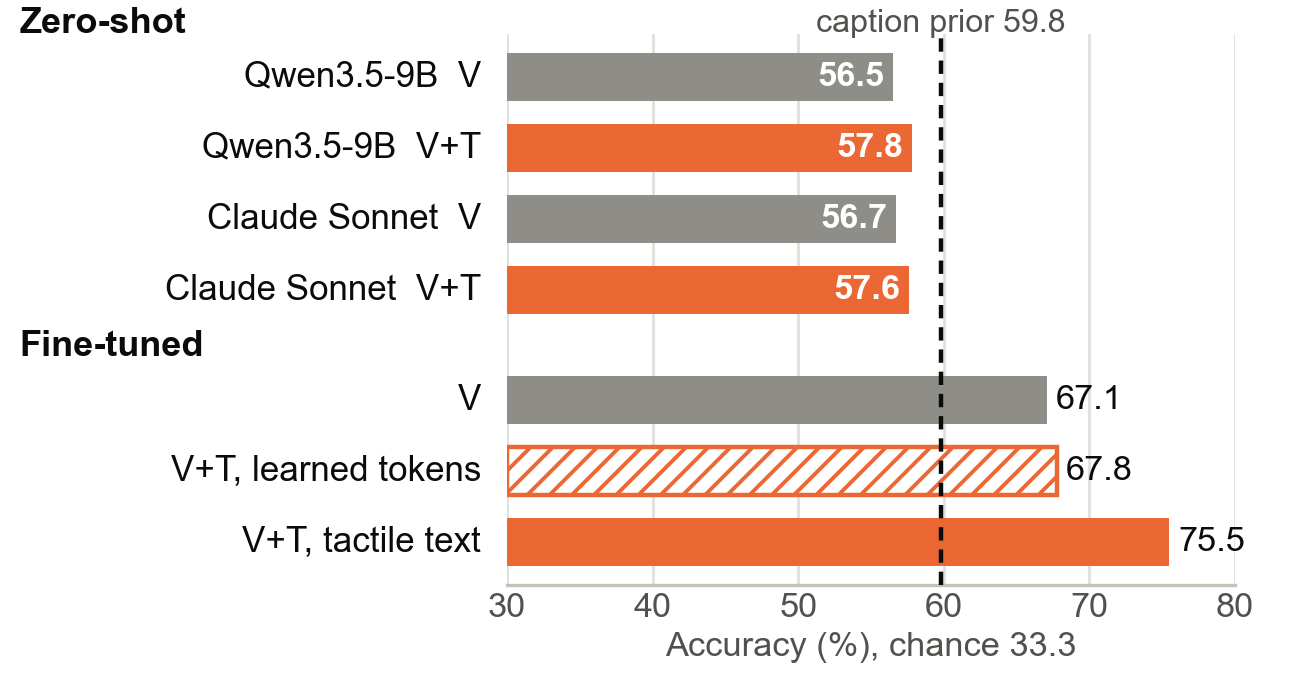}
  \caption{\textbf{Web-scale contact QA} (17{,}544 evaluation questions,
  chance $33.3\%$). No zero-shot model with direct prompting reaches the
  caption-only answer prior (dashed). After fine-tuning, tactile text clears it
  by a wide margin, whereas the learned-token model ignores its tactile input
  (\cref{sec:interface}). Zero-shot rows use direct
  prompting, and fine-tuned rows use Qwen3.5-9B.}
  \label{fig:web}
\end{figure}

\paragraph{Answer prior and shortcuts}
Chance is $33.3\%$, but a bag-of-words classifier on the two captions answers
$59.8\%$ of the items correctly. No zero-shot arm reaches this prior with
direct prompting (\cref{fig:web}): video alone gives $56.5\%$ (Qwen3.5-9B) and $56.7\%$
(Claude Sonnet), and tactile text raises these only to $57.8\%$ and $57.6\%$.
Reasoning-style prompting on a 600-item subsample lifts the strongest arm to
$60.8\%$, $1.0$ point above the prior. Its protocol was set after inspecting
part of this subsample (\cref{app:web}). Pairing events can also introduce
shortcuts: $67.6\%$ of the evaluation items pair events performed with
different hands. On the $4{,}193$ items that pair events of the same hand,
without temporal overlap or ``both'' in either caption, video alone scores
$63.2\%$ and tactile text $72.0\%$, so the tactile gain persists without these
shortcuts.

\section{Conclusion}
\label{sec:conclusion}
\label{sec:limits}

We introduced TouchBench, an egocentric video-tactile benchmark in which
visual answerability is a controlled variable. State-of-the-art VLMs perform near chance on physical properties hidden from
the camera while succeeding, on average, when those properties are visible,
although fine-tuned video models partly recover hidden properties from visual
proxies. Whole-hand touch recovers much of this gap on relative comparisons:
interpretable tactile statistics written into the prompt
require no training, apply to proprietary models, improve relative
comparisons on unseen users and objects and outperform the tactile-token
encoder we trained on the same data.

\paragraph{Limitations}
With 900 clips from five participants, recorded with one right-hand glove, the
benchmark is small, and generalization to unseen users varies across folds:
the fine-tuned TaP model's Tier~B accuracy ranges from $66.0\%$ to $86.0\%$
and its gain over video only from $0.0$ to $12.0$ points. On the dimensions
that contain both tiers, fine-tuned TaP does not improve over video alone.
Tiers follow the capture design and were assigned by one author, without an
agreement study.
Fine-tuning uses one backbone and the learned-token baseline uses one selected
encoder, so a stronger encoder may narrow the gap. Claude models
were sampled once per item, without a dated snapshot. TaP does not
calibrate absolute judgments, the glove's 98\,Hz sampling leaves phone
vibration at chance and the web-scale corpus covers only contact properties.

\paragraph{Broader impact}
Touch-grounded models could make assistive and robotic systems safer around
objects whose state cannot be seen. Conversely, such models inherit the limits
of the sensor and of its summary (here, no vibration above the sampling rate
and no calibrated absolute judgments), so they should not be the only
safeguard where the state of an object matters, for example when handling fragile or hazardous items. 

\section*{LLM Usage Statement}
The authors take full responsibility for all content of this paper, regardless
of any LLM or AI-tool use. We used OpenAI GPT, Anthropic Claude and Google
Gemini models as listed below, giving the highest level of involvement for
each activity.
\emph{Code and implementation (agent):} coding agents carried out multi-step
implementation tasks autonomously, with limited step-by-step human direction.
\emph{Writing and editing, related work and citations, experiment design,
execution and analysis, figures, and data (generate):} the tools produced
drafts of text, citations, analyses, figures and data that the authors
reviewed, edited or approved. This includes the event segmentation and
captions of the web-scale corpus, produced by a Claude model
(\cref{sec:scale}); the answers to those questions are computed from the
tactile signal, not generated by a model. Every bibliography entry was
checked against its published record.
\emph{Research ideas, hypotheses and claims (assist):} the tools only edited
or refined content produced by the authors.
In addition, the full manuscript was given to GPT and Claude models to obtain
reviewer-style feedback. Separately from these uses, Claude Sonnet and Claude
Opus are among the models evaluated in this paper (\cref{sec:exp}).

{\small
\bibliographystyle{ieeenat_fullname}
\bibliography{main}
}

\clearpage
\appendix
\makeatletter
\setlength{\@fptop}{0pt}\setlength{\@fpsep}{12pt plus 4pt}\setlength{\@fpbot}{0pt plus 1fil}
\setlength{\@dblfptop}{0pt}\setlength{\@dblfpsep}{12pt plus 4pt}\setlength{\@dblfpbot}{0pt plus 1fil}
\makeatother
\renewcommand{\topfraction}{0.9}
\renewcommand{\bottomfraction}{0.6}
\renewcommand{\textfraction}{0.1}
\renewcommand{\floatpagefraction}{0.7}
\renewcommand{\dbltopfraction}{0.9}
\renewcommand{\dblfloatpagefraction}{0.7}
\setcounter{topnumber}{4}
\setcounter{dbltopnumber}{3}
\setcounter{totalnumber}{6}
\counterwithin*{table}{section}
\counterwithin*{figure}{section}
\renewcommand{\thetable}{\Alph{section}\arabic{table}}
\renewcommand{\thefigure}{\Alph{section}\arabic{figure}}
\raggedbottom   %
\begin{center}
    \interlinepenalty=10000
    {\Large \bf Appendix \par}
\end{center}
\section{Benchmark details}
\label{app:capture}

\begin{figure*}[tp]
  \centering
  \includegraphics[width=0.95\linewidth]{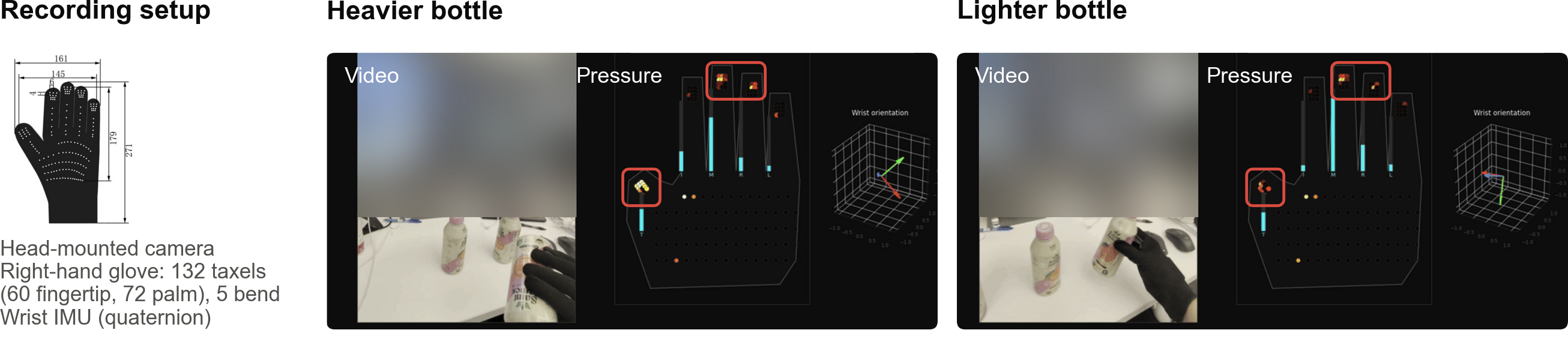}
  \caption{\textbf{Data collection setup.} A head-mounted camera provides the
  egocentric view while a sensorized glove on the right hand records 137
  channels, consisting of 132 pressure taxels over the fingertips and palm and
  5 per-finger flexion channels, together with a wrist inertial measurement
  unit, all time-aligned to the video. The two panels show the synchronized
  viewer at two instants of the same recording. The left bottle is heavier than
  the right, and the difference is visible in the fingertip pressure channels,
  particularly at the thumb, while the two bottles are indistinguishable in the
  video frames. }
  \label{fig:capture}
\end{figure*}

\begin{figure}[tbp]
  \centering
  \includegraphics[width=\linewidth]{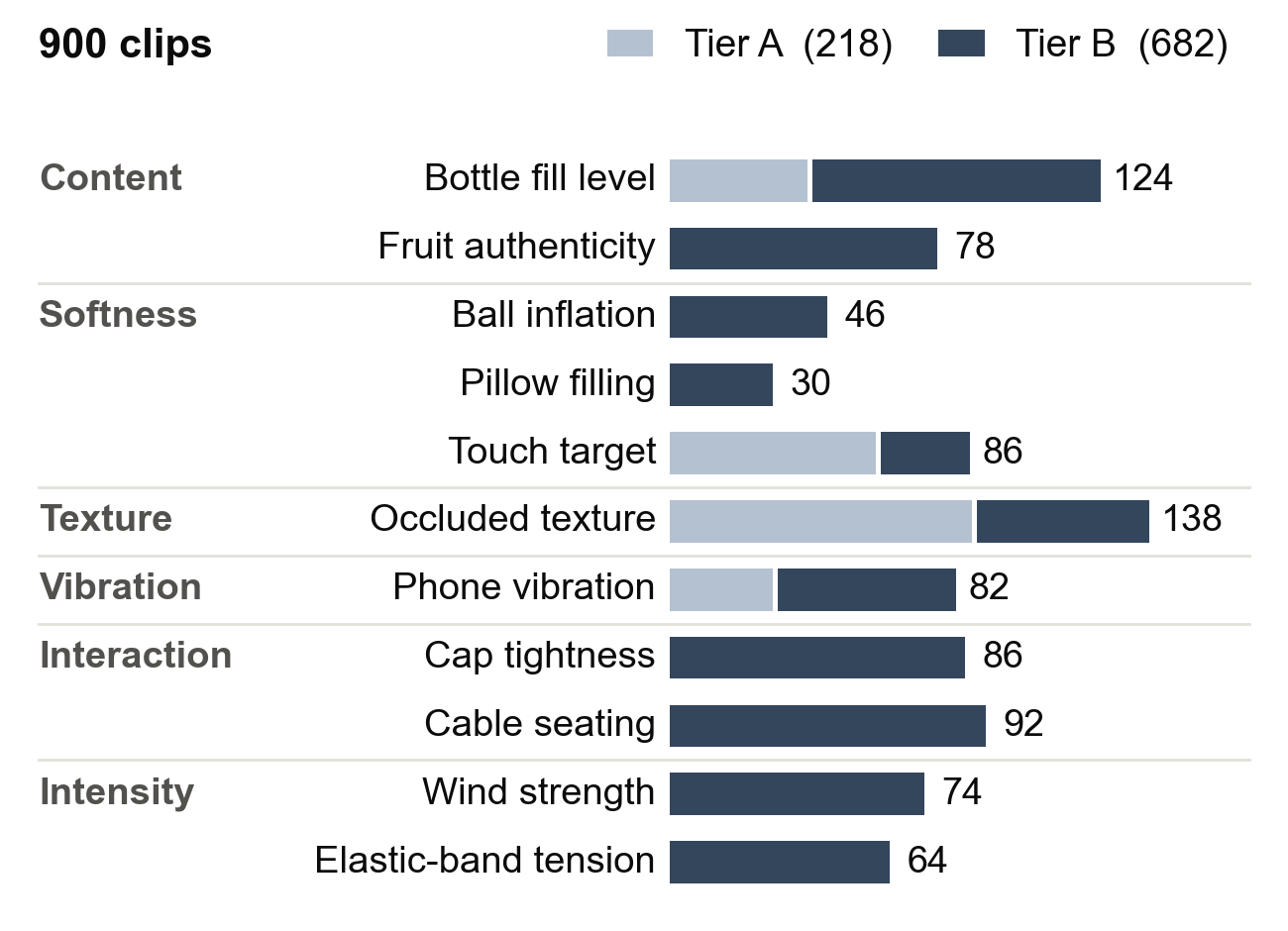}
  \caption{\textbf{Distribution of the 900 clips over the eleven tasks and
  six attribute dimensions, by tier} (218 Tier~A and 682 Tier~B clips). The
  occluded-texture task is scored as two disjoint material pairs, which gives
  twelve scoring dimensions.}
  \label{fig:stats}
\end{figure}

\cref{fig:capture} shows the recording setup and the synchronized viewer used
to verify the alignment between the video and the glove stream.
Participants P1 to P5 contributed 152, 216, 182, 142 and 208 clips, defining
the five leave-one-user-out folds. \cref{fig:stats} shows the distribution
of clips across the eleven tasks and the two tiers.

\paragraph{Data collection}
This work evaluates models on recordings of object manipulation and does not
study the people who made the recordings, so no IRB review was conducted. All
participants signed a consent form before recording.

\paragraph{Sampling rate and synchronization}
The glove streams at about $98$\,Hz: the recording metadata give
$97.7$--$98.3$\,Hz per segment, and timestamp differences give a median of
$98.5$\,Hz across clips. Video and glove are aligned in software. A QR code
shown in the video encodes the Unix epoch in microseconds to fix each
video's start time, which is then subtracted from the glove
timestamps. Three recordings were refit with a unit clock rate to remove
approximately $0.4$--$0.6$\,s of end drift. The residual video-to-glove
offset after synchronization was not measured, because contact onsets in
these grasps cannot be timed precisely in either stream. Each clip is
subsampled to 32 uniformly spaced frames for model input.

\paragraph{Question construction and folds}
Relative pairs are drawn from within a single scoring dimension, so that the
two clips differ in the property being tested rather than in the question
being asked. Every evaluation pair contains two clips of the held-out
participant, whereas training pairs may combine clips of the four training
participants. The training split is capped at 48 pairs per scoring dimension (576 per fold) to keep dimensions balanced. The evaluation split is uncapped, with 606/152, 1{,}040/216, 767/182, 525/142 and 1{,}164/208 relative/absolute items for P1 to P5. No evaluation clip or source segment
appears in a training pair of the same fold. Because each item belongs to the fold
of its held-out participant, pooling the five folds yields the exact
4{,}102 relative and 900 absolute items of the zero-shot rows. Pairs may span visibility conditions within a dimension, such as a fully occluded metal mesh paired with a fully visible metal board. These 922 mixed-tier pairs are reported in \cref{tab:micro}.

\paragraph{Averaging and chance}
Unless noted otherwise, accuracy is the unweighted mean of capture-subcategory
accuracies. Each relative item is attributed to the task and subcategory of its
first clip. The two texture dimensions share their subcategories for this
average, yielding six Tier~A and sixteen Tier~B subcategories
(\cref{tab:subcategories}). Every relative question has two options. Absolute questions have two options except for the elastic-band count, which has four. Chance accuracy for absolute questions is the mean of $1/k$ over subcategories, yielding $50.0\%$ on Tier~A and $48.4\%$ on Tier~B.

\paragraph{Tier assignment}
As stated in \cref{sec:tiers}, tiers were assigned once per
capture subcategory rather than per item: one author inspected the capture
conditions and marked a subcategory Tier~A only when its target property could
be determined from the pixels without using object identity or another
statistical proxy. \cref{tab:subcategories} gives the complete mapping.
\begin{table*}[tp]
\centering\small
\setlength{\tabcolsep}{4pt}
\begin{tabular}{@{}llcrrrrrr@{}}
\toprule
& & & & & \multicolumn{4}{c}{Zero-shot \VO{} accuracy} \\
\cmidrule(lr){6-9}
Task & Subcategory & Tier & Clips & Rel.\ pairs & InternVL3-8B & Qwen3.5-9B & Claude Sonnet & Claude Opus \\
\midrule
Elastic-band tension & -- & B & 64 & 348 & 56.0 & 73.3 & 58.9 & 59.5 \\
\addlinespace[2pt]
Bottle fill level & opaque & B & 38 & 159 & 56.0 & 49.1 & 47.8 & 47.8 \\
 & semi-transparent & B & 46 & 213 & 56.3 & 49.3 & 45.1 & 47.4 \\
 & transparent & A & 40 & 84 & 83.3 & 66.7 & 59.5 & 65.5 \\
\addlinespace[2pt]
Cable seating & fully occluded & B & 26 & 117 & 53.8 & 66.7 & 64.1 & 59.0 \\
 & fully visible & B & 32 & 160 & 44.4 & 57.5 & 61.2 & 47.5 \\
 & partial & B & 34 & 169 & 48.5 & 47.9 & 56.8 & 43.2 \\
\addlinespace[2pt]
Cap tightness & -- & B & 86 & 387 & 56.1 & 56.8 & 51.2 & 50.1 \\
\addlinespace[2pt]
Fruit authenticity & apple & B & 40 & 177 & 37.9 & 52.0 & 54.8 & 55.9 \\
 & banana & B & 38 & 182 & 47.3 & 48.4 & 49.5 & 45.1 \\
\addlinespace[2pt]
Touch target & fully occluded & B & 26 & 37 & 37.8 & 29.7 & 45.9 & 48.6 \\
 & fully visible & A & 30 & 94 & 92.6 & 90.4 & 74.5 & 57.4 \\
 & partial & A & 30 & 94 & 85.1 & 77.7 & 74.5 & 52.1 \\
\addlinespace[2pt]
Phone vibration & face screen & A & 30 & 45 & 75.6 & 100.0 & 44.4 & 100.0 \\
 & screen back & B & 26 & 74 & 52.7 & 41.9 & 47.3 & 37.8 \\
 & screen side & B & 26 & 64 & 50.0 & 57.8 & 51.6 & 62.5 \\
\addlinespace[2pt]
Ball inflation & ball & B & 46 & 121 & 67.8 & 47.1 & 44.6 & 54.5 \\
\addlinespace[2pt]
Pillow filling & pillow & B & 30 & 59 & 42.4 & 40.7 & 45.8 & 25.4 \\
\addlinespace[2pt]
Texture, material & fully occluded & B & 20 & 24 & 50.0 & 41.7 & 58.3 & 54.2 \\
 & fully visible & A & 18 & 32 & 93.8 & 100.0 & 87.5 & 78.1 \\
 & partial & A & 20 & 53 & 96.2 & 100.0 & 84.9 & 75.5 \\
\addlinespace[2pt]
Texture, structure & fully occluded & B & 30 & 49 & 24.5 & 28.6 & 28.6 & 24.5 \\
 & fully visible & A & 24 & 66 & 12.1 & 25.8 & 43.9 & 13.6 \\
 & partial & A & 26 & 75 & 14.7 & 33.3 & 29.3 & 33.3 \\
\addlinespace[2pt]
Wind strength & -- & B & 74 & 297 & 60.9 & 45.5 & 50.2 & 60.9 \\
\bottomrule
\end{tabular}
\caption{\textbf{Tier assignment and counts per capture subcategory.} Each clip has one absolute question. Cable seating is Tier~B in every capture condition, because whether the plug is fully seated cannot be seen even when the connector is in view. Relative counts and video-only accuracies use same-tier pairs, attributed to the first clip when subcategories differ; this convention partitions the 543 Tier~A and 2,637 Tier~B pairs without double-counting. Tiers are the corrected assignment after the audit below (semi-transparent bottle moved to Tier~B). Columns show zero-shot relative video-only accuracy (\%).}
\label{tab:subcategories}
\end{table*}

\paragraph{Label and tier audit}
\label{app:audit}
We audited three groups whose zero-shot scores were suspicious, with the
texture material pair as a control, by answering each item from its frames,
question and options alone and then comparing with the gold answer.
(a) \emph{Structure pair, fully visible and partial (Tier~A)}, where all
video-only models score far below chance: 8 of 10 items could be judged from
the frames (the other two show both surfaces mostly covered), and all 8 agree
with the gold answer. The question names the target explicitly (``Which surface is more mesh-like / open (metal mesh rather than a solid metal board)?''), so neither the labels nor the wording is inverted. Rather, the zero-shot models systematically pick the solid board, whereas the fine-tuned video model does not ($99.3\%$). (b) \emph{Semi-transparent bottle (Tier~A)}: the amber
bottle shows no liquid line or fill difference in any inspected frame. The 5 of 10 pairs that compare two amber bottles could not be judged, and the other 5 were answerable only from the transparent bottle in the pair. The amber bottles in the unseen-object split look identical. This subcategory is
therefore reassigned to Tier~B. (c) \emph{Elastic bands (Tier~B)}: in 20 relative pairs, 16 could be judged with low confidence from the thickness of the band bundle stretched between the hands, and 13 of these agree with the gold answer. In 10 absolute items, the band count was read correctly once. The labels are correct and the tier is kept, because the count itself is not visible. The bundle thickness is a partial visual proxy, which is consistent with the above-chance zero-shot scores. (d) \emph{Control (material pair,
Tier~A)}: 10 of 10 items could be judged, all agreeing with the gold answer.
No gold label was changed. After this reassignment, Tier~A has six capture subcategories and Tier~B has sixteen. There are 543 Tier~A, 2{,}637 Tier~B and 922 mixed-tier relative pairs, and 218 Tier~A and 682 Tier~B absolute items. All tables in this paper use the corrected tiers.

\section{Tactile descriptors}
\label{app:features}

\begin{figure}[tbp]
  \centering
  \includegraphics[width=\linewidth]{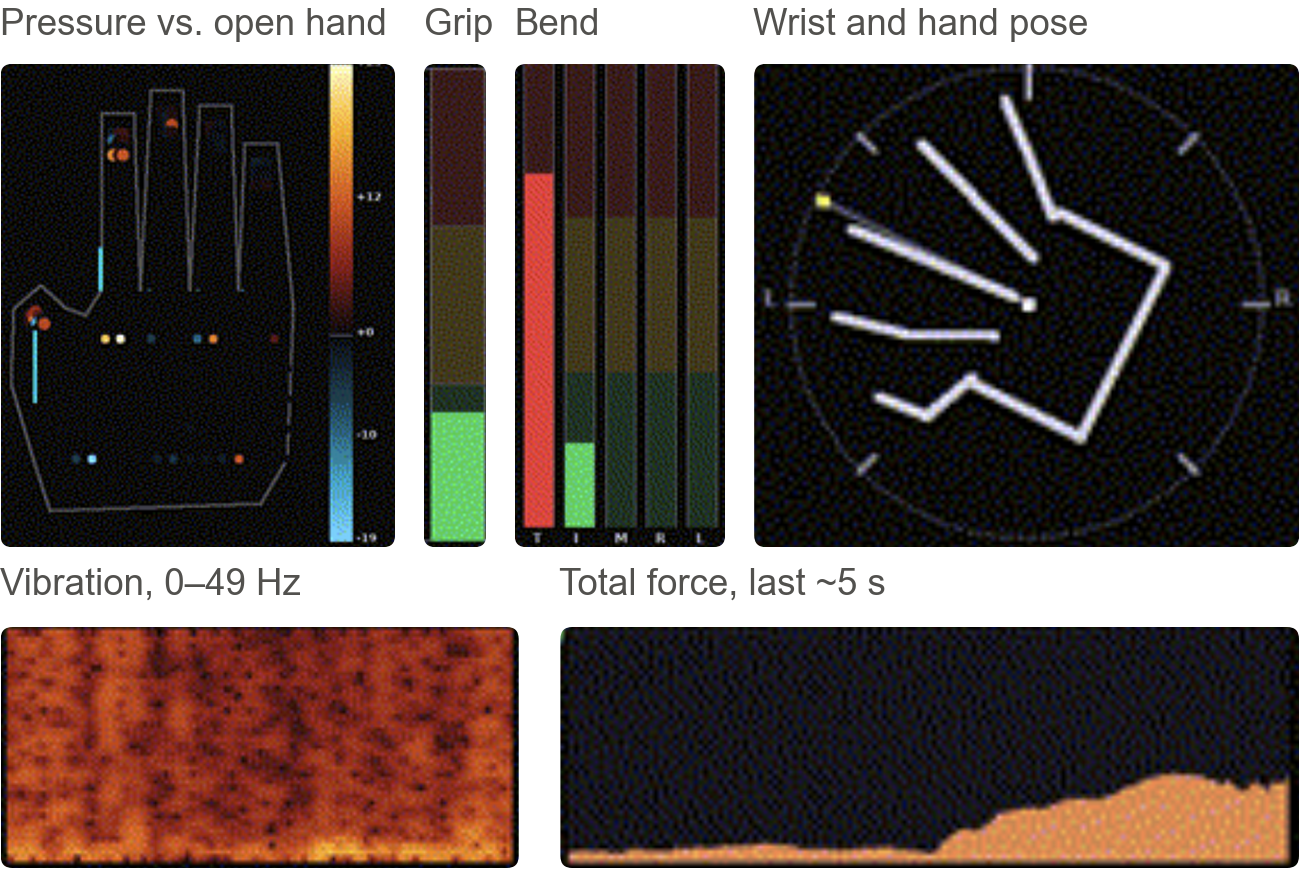}
  \caption{\textbf{The tactile stream that Tactile-as-Prompt summarizes.} A
  single grasp is shown: per-taxel pressure on a hand outline (left, signed
  relative to an open-hand zero), fingertip grip force, wrist twist and hand
  pose, a $0$--$49$\,Hz vibration spectrogram, total force over the last few
  seconds and per-finger bend. The 34 statistics in \cref{tab:features} summarize these channels.}
  \label{fig:coviz}
\end{figure}

\cref{tab:features} defines the 34 statistics that Tactile-as-Prompt
serializes, grouped into four blocks. These four blocks serve as the
units for the descriptor-group ablation in \cref{app:results}. \cref{fig:coviz} shows the raw channels that these statistics
summarize.

\begin{table*}[tp]
  \centering\small
  \setlength{\tabcolsep}{5pt}
  \renewcommand{\arraystretch}{1.04}
  \begin{tabularx}{\linewidth}{@{}l >{\raggedright\arraybackslash}X@{}}
    \toprule
    Feature & Definition (units where determinable) \\
    \midrule
    \multicolumn{2}{@{}l}{\textbf{Force} \textnormal{(14 statistics)}} \\
    \addlinespace[1pt]
    \texttt{f\_peak} & Max over frames of sum of rectified channels; whole-hand signal peak (mixed units). \\
    \texttt{f\_p95} & 95th percentile of rectified-channel sum over frames; robust peak (mixed units). \\
    \texttt{f\_mean} & Mean rectified-channel sum over frames (mixed units). \\
    \texttt{f\_std} & Temporal standard deviation of the rectified-channel sum (mixed units). \\
    \texttt{max\_single} & Largest rectified value among all 137 channels (mixed units). \\
    \texttt{finger0} & Peak summed pressure of the thumb pad (p.u.). \\
    \texttt{finger1} & Peak summed pressure of the index pad (p.u.). \\
    \texttt{finger2} & Peak summed pressure of the middle pad (p.u.). \\
    \texttt{finger3} & Peak summed pressure of the ring pad (p.u.). \\
    \texttt{finger4} & Peak summed pressure of the little-finger pad (p.u.). \\
    \texttt{fingertip\_total} & Peak summed pressure over all fingertip taxels (p.u.). \\
    \texttt{palm\_force} & Peak summed pressure over all palm taxels (p.u.). \\
    \texttt{fp\_ratio} & Fingertip peak divided by palm peak; fingertip-to-palm force ratio, clamped to 99.9 when serialized (dimensionless). \\
    \texttt{flex\_sum} & Sum over the five per-finger bend channels of their signed (unrectified) per-channel time mean; overall flexion (bend units). \\
    \midrule
    \multicolumn{2}{@{}l}{\textbf{Spatial} \textnormal{(8 statistics)}} \\
    \addlinespace[1pt]
    \texttt{active\_count} & Number of channels on the peak frame above $\max(2.0,\,0.1\times\text{frame max})$; active-channel count (channels). \\
    \texttt{gini} & Gini concentration of active-channel values ($0$ uniform to $1$ concentrated; dimensionless). \\
    \texttt{top1\_frac} & Strongest active channel divided by the sum of active channels (dimensionless). \\
    \texttt{top3\_frac} & Sum of the three strongest active channels divided by the sum of active channels (dimensionless). \\
    \texttt{spatial\_cv} & Standard deviation divided by mean of active-channel values; spatial coefficient of variation (dimensionless). \\
    \texttt{n\_peaks} & Count of upward crossings of rectified-channel sum above $0.5\times$\,\texttt{f\_peak}; number of signal pulses (count). \\
    \texttt{f\_slope} & Maximum single-frame increase in rectified-channel sum; fastest onset (mixed units per frame). \\
    \texttt{peak\_time\_frac} & Peak-sum frame index divided by $(T-1)$; normalized time of the peak, $0$ to $1$ (dimensionless). \\
    \midrule
    \multicolumn{2}{@{}l}{\textbf{Spectral} \textnormal{(10 statistics)}} \\
    \addlinespace[1pt]
    \texttt{spec\_b0816\_mean} & Mean over active channels of the Welch sum of PSD bins, 8--16\,Hz (PSD). \\
    \texttt{spec\_b0816\_max} & Max over active channels of the sum of PSD bins, 8--16\,Hz (PSD). \\
    \texttt{spec\_b1632\_mean} & Mean sum of PSD bins, 16--32\,Hz (PSD). \\
    \texttt{spec\_b1632\_max} & Max sum of PSD bins, 16--32\,Hz (PSD). \\
    \texttt{spec\_b3249\_mean} & Mean sum of PSD bins, 32--49\,Hz (PSD). \\
    \texttt{spec\_b3249\_max} & Max sum of PSD bins, 32--49\,Hz (PSD). \\
    \texttt{spec\_b4449\_mean} & Mean sum of PSD bins, 44--49\,Hz; narrow high band (PSD). \\
    \texttt{spec\_b4449\_max} & Max sum of PSD bins, 44--49\,Hz (PSD). \\
    \texttt{spec\_centroid} & Mean over active channels of the PSD spectral centroid $\sum fP/\sum P$ (Hz). \\
    \texttt{hf\_diff\_rms} & Mean absolute frame-to-frame change of the unrectified values of the spectrally active channels; frame-to-frame jitter (mixed units). \\
    \midrule
    \multicolumn{2}{@{}l}{\textbf{Wrist} \textnormal{(2 statistics)}} \\
    \addlinespace[1pt]
    \texttt{wrist\_jitter} & Mean absolute first difference across the four wrist-quaternion components (dimensionless). \\
    \texttt{wrist\_tilt\_range} & Range (max minus min) of wrist-quaternion component $q_1$; component excursion (dimensionless), not an angle. \\
    \bottomrule
  \end{tabularx}
  \caption{\textbf{The 34 tactile statistics} computed per clip from the
    baseline-subtracted glove stream (60 fingertip taxels, 5 per-finger bend
    channels and 72 palm taxels) and the wrist quaternion, organized into four
    descriptor groups. ``p.u.'' denotes sensor pressure units, which are not calibrated to
    physical force. Values enter the prompt in raw units without
    standardization.}
  \label{tab:features}
\end{table*}

\paragraph{Implementation notes}
The implementation rectifies all 137 channels for the whole-hand totals and
spatial statistics, including the five bend channels, so those totals are
mixed sensor-unit summaries rather than calibrated force. In contrast, finger and palm
subsets contain pressure channels only. Spectral activity selects channels
with a peak above 15 and a value above 10 in at least 5\% of frames,
including bend channels when they qualify. Welch powers sum the bins with
$l\leq f<h$, using a sample rate estimated from the timestamps. At the typical
rate of about 98.5\,Hz, the Nyquist frequency lies just above 49\,Hz, so the
upper bands end at the last bin below 49\,Hz. The tree baseline reads the same
34 features, standardized with training-participant means and standard
deviations, an affine transform that leaves tree splits unchanged.

\paragraph{Serialization}
The statistics are formatted in a fixed order (force, spatial, spectral,
wrist) as semicolon-separated clauses with plain-English names and units.
\cref{app:prompts} provides the exact template and rounding rules. Prompt values keep
the extracted sensor scales: they are rounded, the fingertip-to-palm ratio is
clamped to $99.9$ and no $z$-scoring is applied.

\section{Models and training}
\label{app:training}

\paragraph{Zero-shot evaluation}
The open-weight checkpoints are \texttt{OpenGVLab/InternVL3-8B} and
\texttt{Qwen/Qwen3.5-9B}. Question and option text is fixed across input
conditions, while the system preamble describes the supplied modalities
(\cref{app:prompts}). The open models receive 32 frames per clip and are
scored without generation: thinking is disabled and the option letter with the
largest next-token logit among the offered options is selected. Every question
has two options (A or B) except the absolute elastic-band count, which has
four (A to D). The proprietary models were queried on 14 October 2026
through a non-interactive command-line interface with all tools disabled.
Every response reports the model as \texttt{claude-sonnet-5} or
\texttt{claude-opus-5}. The 32 frames of each clip are placed in the user
message as two contact-sheet images, followed by the same instruction,
question and options. A single letter is requested at the default
sampling settings because this interface exposes no temperature control. Only
outputs consisting of an option letter are scored. All zero-shot responses on
TouchBench are valid. Open-model and LOUO entries report one deterministic run
per arm, and proprietary entries report one sample per item.

\paragraph{Fine-tuning}
All fine-tuning experiments use Qwen3.5-9B. Both tactile interfaces fine-tune
it on the same training split, prompt scaffold and schedule. The backbone is
frozen and LoRA ($r{=}32$, $\alpha{=}64$, dropout $0.05$) is applied to the
attention and MLP projections. AdamW uses a peak learning rate of
$2\times10^{-4}$ with a linear schedule, 3\% warm-up, bfloat16 and an
effective batch size of 16. Relative and absolute models train for five and
ten epochs, respectively, on one H100 per run.

\paragraph{Learned-token encoder}
The learned interface is a 1.79M-parameter temporal encoder: four stride-2
1-D convolution blocks map the 137 channels to 256 hidden channels, linear
interpolation produces $k{=}32$ time-aligned tokens, and a linear projection,
layer normalization and learned gate map them to the 4{,}096-dimensional
language space. It is initialized by symmetric frame-level InfoNCE against
frozen visual features, with an additional mean-pooled clip term, on the
50-minute free-form set and the 20-hour public corpus (\cref{fig:pretrain}).
The $21\times21$ pressure grid of the public corpus is mapped onto a shared hand
canvas, assigned to the 132 target pressure positions by nearest neighbor with
collision averaging and augmented with five bend channels. Each arm adapts
the language model with 58{,}195{,}968 LoRA parameters. The learned-token arm
additionally updates a 1{,}792{,}129-parameter encoder and a
1{,}048{,}833-parameter conditional gate (61{,}036{,}930 trainable parameters
in total), whereas TaP adds none. Our learned-token experiments varied fusion, gating,
normalization and pre-training, so the reported configuration is a reference
rather than a tuned optimum.

\paragraph{Unseen-object split}
The 225 unseen-object clips were recorded by one of the five participants with
object instances that do not appear in the benchmark. The model is trained on
the benchmark clips of the other four participants and evaluated on this held-out set, so both the objects and the user are unseen. The relative set contains 906 pairs (95 Tier~A, 653 Tier~B and 158
mixed-tier) over nine scoring dimensions, since no touch-target contrast is
available. The absolute set contains 225 clips over ten scoring dimensions,
excluding cap tightness and pillow filling. Seeds 42, 43 and 44 change model
initialization and training-example order, while evaluation is deterministic.
The \VO{} and \VT{} results average all three seeds, and the \TO{} results average the
first two.

\section{Additional results}
\label{app:results}

This section expands on the results of the main text, breaking down the main table by tier and by item, and detailing the controls and their uncertainty, per-dimension accuracy, the unseen-object split, the tactile interface and free-form captioning.

\subsection{Main results by tier}
\begin{figure*}[tp]
  \centering
  \includegraphics[width=\linewidth]{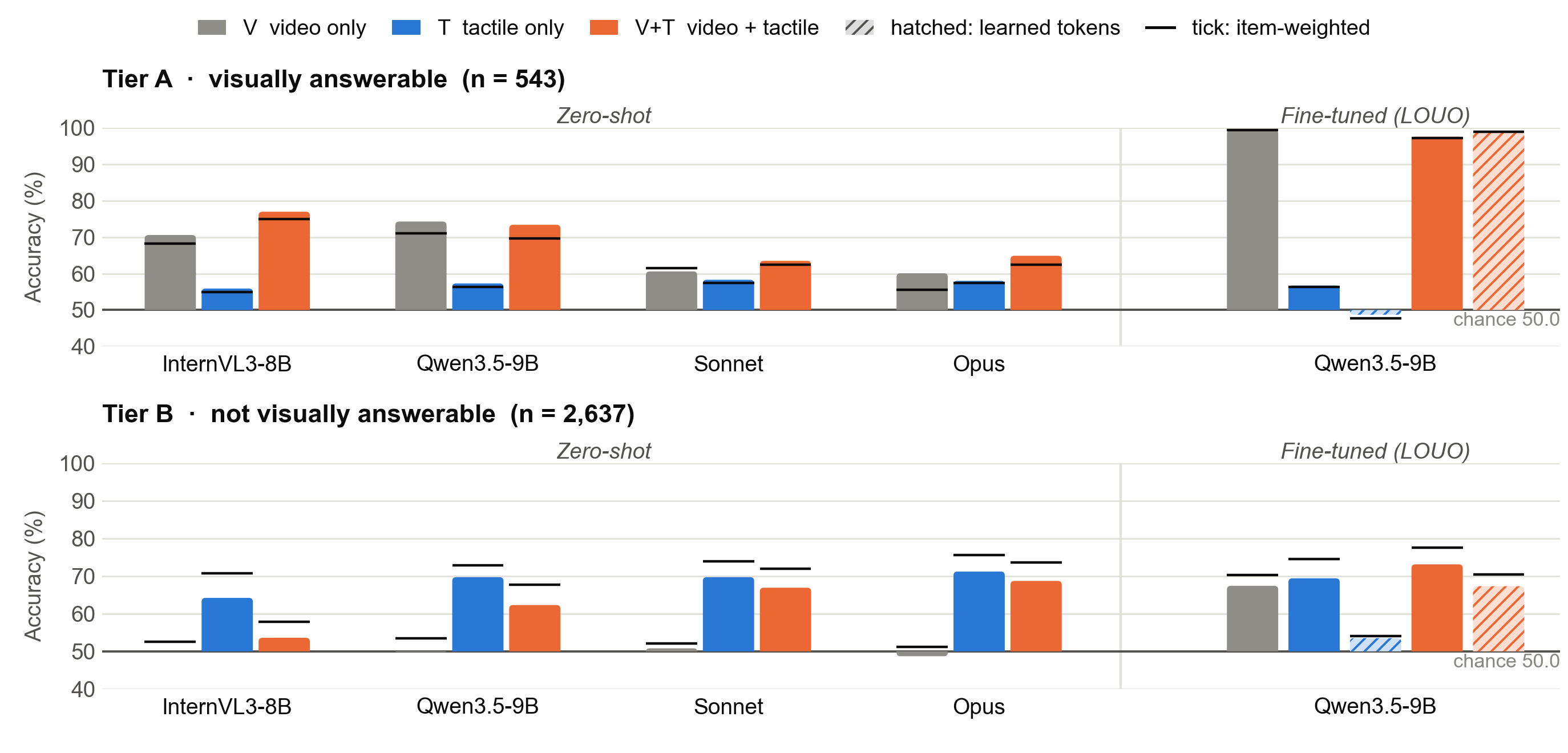}
  \caption{\textbf{Relative (two-clip) comparison for all four models and the
  fine-tuned model, by tier.} Bar height is the macro-average over capture subcategories, as in \cref{tab:main}, and the tick marks the item-weighted (micro) average.
  Video only scores well
  on Tier~A, which confirms the control. On Tier~B, every video-only arm sits
  near the $50.0$ chance line, whereas the tactile arms lift accuracy above it. Hatched
  bars denote the learned-token variant. The proprietary arms use the re-run with the frames supplied in the prompt. The earlier agent-harness runs are compared in \cref{tab:claude-harness}.}
  \label{fig:res_rel}
\end{figure*}

\begin{figure*}[tp]
  \centering
  \includegraphics[width=\linewidth]{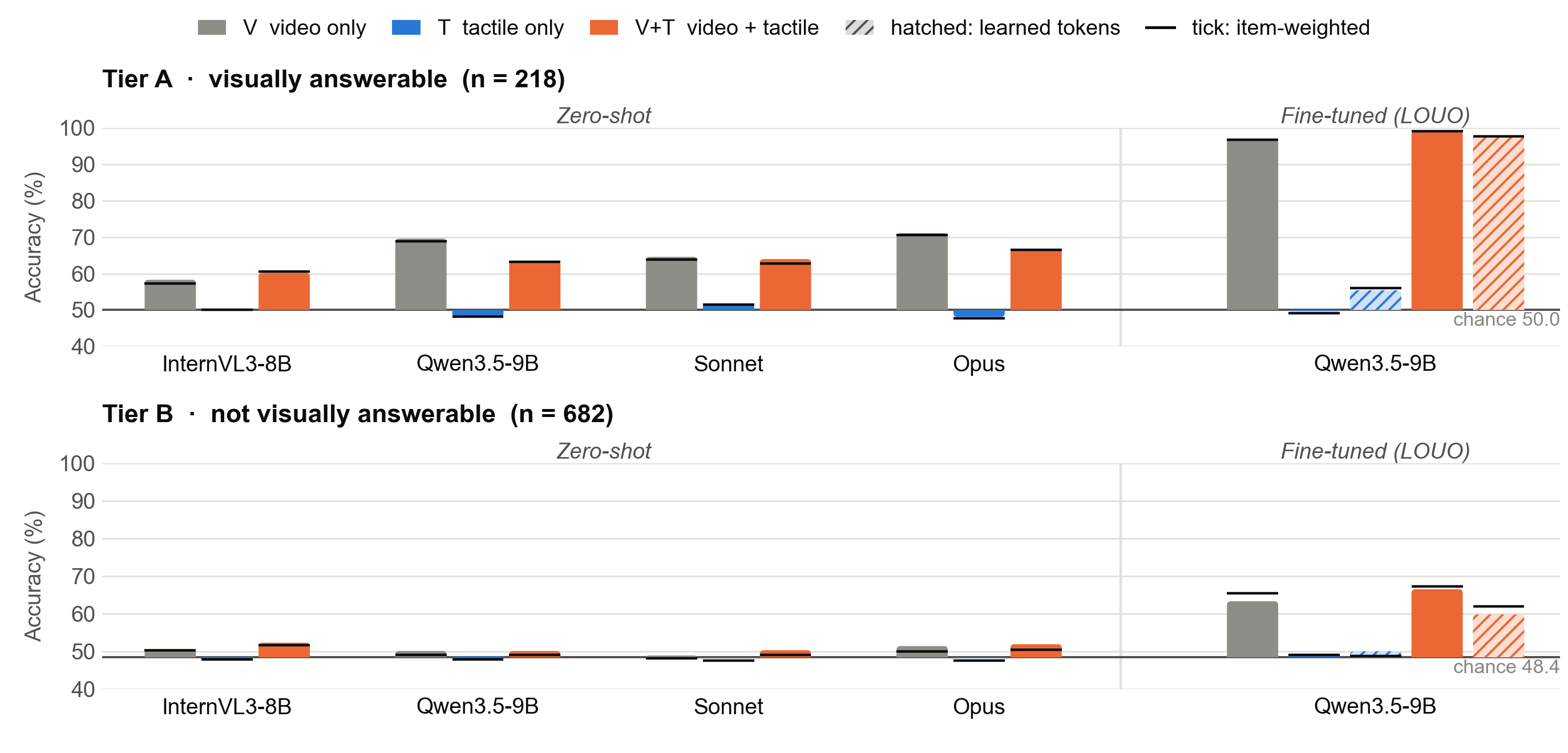}
  \caption{\textbf{Absolute (single-clip) results for the same models and
  tiers.} In contrast to \cref{fig:res_rel}, adding tactile input does not
  help systematically: the model uses touch far more effectively to compare two
  clips than to calibrate a single absolute judgment. Bars and ticks as in
  \cref{fig:res_rel}. Chance is $50.0$ on Tier~A and $48.4$ on Tier~B, because
  the four-option band count is a Tier~B task.}
  \label{fig:res_abs}
\end{figure*}

\cref{fig:res_rel,fig:res_abs} plot \cref{tab:main} for every arm,
with the subcategory macro-average as bars and the item-weighted average as
ticks. \cref{tab:micro} lists the item-weighted values, including the 922
mixed-tier pairs that the tier-specific results exclude. The item-weighted
averages preserve the pattern of the main text. On mixed-tier pairs, which
can partly be answered from the visible clip, fine-tuned \VT{} TaP reaches
$92.7\%$ against $87.9\%$ for video alone. \cref{tab:claude-harness} compares
the proprietary runs of the main text with an earlier harness in which the
models could skip reading the frames.

\begin{table*}[tp]
\centering\small
\setlength{\tabcolsep}{3pt}
\begin{tabular}{@{}lrrrrrr@{}}
\toprule
Model / input & Rel. A & Rel. B & Rel. all & Rel. mixed & Abs. A & Abs. B \\
\midrule
Qwen \VO & 71.1 & 53.4 & 55.6 & 52.6 & 68.8 & 49.1 \\
Qwen \VT & 69.6 & 67.7 & 66.3 & 60.4 & 63.3 & 49.1 \\
Qwen \TO & 56.4 & 72.8 & 68.6 & 63.8 & 48.2 & 47.8 \\
InternVL \VO & 68.3 & 52.6 & 55.2 & 55.1 & 57.3 & 50.3 \\
InternVL \VT & 75.0 & 57.8 & 59.8 & 56.6 & 60.6 & 51.6 \\
InternVL \TO & 54.9 & 70.8 & 65.4 & 55.9 & 50.0 & 47.9 \\
Sonnet \VO & 61.5 & 52.1 & 53.1 & 51.2 & 63.8 & 48.1 \\
Sonnet \VT & 62.4 & 72.0 & 67.6 & 58.1 & 62.8 & 49.0 \\
Sonnet \TO & 57.5 & 73.9 & 68.8 & 61.0 & 51.4 & 47.5 \\
Opus \VO & 55.6 & 51.2 & 51.0 & 47.7 & 70.6 & 49.9 \\
Opus \VT & 62.4 & 73.6 & 69.3 & 60.8 & 66.5 & 50.4 \\
Opus \TO & 57.5 & 75.6 & 69.7 & 60.1 & 47.7 & 47.5 \\
Zero-shot $\VO$ mean & 64.1 & 52.3 & 53.7 & 51.7 & 65.1 & 49.3 \\
Fine-tuned \VO & 99.4 & 70.2 & 78.0 & 87.9 & 96.8 & 65.5 \\
Fine-tuned \VT{} (TaP) & 97.2 & 77.5 & 83.5 & 92.7 & 99.1 & 67.2 \\
Fine-tuned \VT{} (tokens) & 98.9 & 70.5 & 78.0 & 86.9 & 97.7 & 62.0 \\
Fine-tuned \TO{} (TaP) & 56.4 & 74.5 & 68.4 & 58.2 & 49.1 & 49.1 \\
Fine-tuned \TO{} (tokens) & 47.7 & 54.0 & 53.0 & 53.6 & 56.0 & 48.7 \\
\bottomrule
\end{tabular}
\caption{\textbf{Item-weighted accuracy} (\%). The relative all column includes 4,102 pairs and the mixed column 922 pairs. The two same-tier columns contain 543 and 2,637 pairs. Absolute columns contain 218 and 682 items.}
\label{tab:micro}
\end{table*}

\begin{table*}[tp]
\centering\small
\setlength{\tabcolsep}{3pt}
\begin{tabular}{@{}lllrrrrrr@{}}
\toprule
& & & \multicolumn{2}{c}{Relative} & \multicolumn{2}{c}{Absolute} & & \\
\cmidrule(lr){4-5}\cmidrule(lr){6-7}
Model & Harness & Input & Tier~A & Tier~B & Tier~A & Tier~B & Unparsed (rel. / abs.) & No image read (rel. / abs.) \\
\midrule
Sonnet & agent & \VO & 66.9 & 52.1 & 50.9 & 49.2 & 4 / 0 & 0 / 697 \\
Sonnet & agent & \TO & 51.1 & 60.1 & 48.3 & 48.6 & 0 / 0 & -- \\
Sonnet & agent & \VT & 54.1 & 62.4 & 48.0 & 49.1 & 0 / 0 & 3{,}769 / 759 \\
Sonnet & CLI & \VO & 60.6 & 50.8 & 64.6 & 48.9 & 0 / 0 & -- \\
Sonnet & CLI & \TO & 58.3 & 69.7 & 51.8 & 48.3 & 0 / 0 & -- \\
Sonnet & CLI & \VT & 63.5 & 66.9 & 64.0 & 50.3 & 0 / 0 & -- \\
\addlinespace[2pt]
Opus & agent & \VO & 58.3 & 51.2 & 63.1 & 52.9 & 0 / 0 & 0 / 0 \\
Opus & agent & \TO & 59.2 & 71.7 & 50.0 & 48.4 & 0 / 0 & -- \\
Opus & agent & \VT & 56.0 & 69.6 & 59.6 & 50.5 & 0 / 0 & 3{,}829 / 223 \\
Opus & CLI & \VO & 60.1 & 48.7 & 71.1 & 51.4 & 0 / 0 & -- \\
Opus & CLI & \TO & 58.0 & 71.2 & 48.1 & 48.2 & 0 / 0 & -- \\
Opus & CLI & \VT & 64.9 & 68.7 & 66.8 & 51.9 & 0 / 0 & -- \\
\bottomrule
\end{tabular}
\caption{\textbf{Proprietary models under two harnesses:} the original agent harness and the re-run used in the main text (macro accuracy, \%, corrected tiers). In the agent runs, each model received the contact sheets as files through a file-reading tool that it could choose not to call. The models were called through the aliases \texttt{sonnet} and \texttt{claude-opus-5}, the resolved snapshots and query dates were not recorded and the archived outputs carry file dates of 14 August 2026. ``No image read'' counts multimodal responses with no logged file read in the tool-call traces. In the re-run, the sheets are placed in the user message as images and no tools are available. Every response reports the model as \texttt{claude-sonnet-5} or \texttt{claude-opus-5}, and all queries were made on 14 October 2026.}
\label{tab:claude-harness}
\end{table*}

\subsection{Controls and uncertainty}
\cref{tab:shared-dims} restricts the relative comparison to the five scoring
dimensions that occur in both tiers. Every Tier~A subcategory belongs to these
dimensions, so only the Tier~B columns change: video-only accuracy drops to
$45.9\%$ on average, and fine-tuned TaP and video alone are on par
($62.2\%$ and $62.9\%$, macro). \cref{tab:swap} evaluates every relative pair in
both clip orders. With video alone, the zero-shot models pick the same clip in
both orders for only $32.2$--$48.8\%$ of Tier~B pairs, against $75.1\%$ for
the fine-tuned video model and $84.6\%$ for fine-tuned TaP. \cref{tab:stats}
gives bootstrap intervals and McNemar tests for the comparisons of the main
text. \cref{tab:app-controls} replaces or removes the tactile line at inference:
shuffling lowers Tier~B accuracy by $11.3$ to $19.7$ points for every arm
except zero-shot InternVL with video ($2.3$). \cref{tab:folds} splits the
fine-tuned results by held-out participant: the gain of TaP over video on
Tier~B is $8.9$, $12.0$, $0.3$, $2.9$ and $0.0$ points for P1 to P5.

\begin{table}[tbp]
\centering\small
\setlength{\tabcolsep}{3pt}
\begin{tabular}{@{}lrrrr@{}}
\toprule
& \multicolumn{2}{c}{Tier~A} & \multicolumn{2}{c}{Tier~B} \\
\cmidrule(lr){2-3}\cmidrule(lr){4-5}
Model / input & Macro & Micro & Macro & Micro \\
\midrule
Qwen \VO & 74.3 & 71.1 & 43.4 & 46.1 \\
Qwen \TO & 57.3 & 56.4 & 65.1 & 72.6 \\
Qwen \VT & 73.4 & 69.6 & 53.7 & 59.2 \\
InternVL \VO & 70.6 & 68.3 & 47.6 & 51.3 \\
InternVL \TO & 55.9 & 54.9 & 59.1 & 69.4 \\
InternVL \VT & 77.0 & 75.0 & 50.6 & 56.1 \\
Sonnet \VO & 60.6 & 61.5 & 46.0 & 46.0 \\
Sonnet \TO & 58.3 & 57.5 & 63.7 & 72.1 \\
Sonnet \VT & 63.5 & 62.4 & 60.2 & 69.4 \\
Opus \VO & 60.1 & 55.6 & 46.4 & 46.5 \\
Opus \TO & 58.0 & 57.5 & 66.8 & 75.6 \\
Opus \VT & 64.9 & 62.4 & 65.3 & 73.5 \\
Fine-tuned \VO & 99.5 & 99.4 & 62.9 & 65.3 \\
Fine-tuned \VT{} (TaP) & 97.6 & 97.2 & 62.2 & 70.6 \\
\bottomrule
\end{tabular}
\caption{\textbf{Relative accuracy on the shared dimensions} (\%), restricted to the five scoring dimensions present in both tiers (bottle, glove, phone and the two texture pairs). Macro is the capture-subcategory mean as in \cref{tab:main}. $n=543$ Tier~A pairs (6 subcategories) and $n=620$ Tier~B pairs (6 subcategories). Every Tier~A subcategory belongs to these five dimensions, so the Tier~A columns equal the unrestricted ones.}
\label{tab:shared-dims}
\end{table}

\begin{table*}[tp]
\centering\small
\setlength{\tabcolsep}{2.5pt}
\begin{tabular}{@{}lrrrrrrrrrrrr@{}}
\toprule
& \multicolumn{6}{c}{Tier~A ($n=543$)} & \multicolumn{6}{c}{Tier~B ($n=2{,}637$)} \\
\cmidrule(lr){2-7}\cmidrule(lr){8-13}
Model / input & orig. & swap. & avg. & same clip & B (orig.) & B (swap.) & orig. & swap. & avg. & same clip & B (orig.) & B (swap.) \\
\midrule
Qwen \VO & 74.2 & 76.7 & 75.5 & 72.6 & 57.6 & 59.5 & 49.9 & 51.3 & 50.6 & 47.4 & 71.9 & 71.6 \\
Qwen \TO & 58.2 & 58.3 & 58.3 & 79.9 & 51.0 & 50.6 & 69.8 & 71.0 & 70.4 & 82.4 & 52.3 & 54.0 \\
Qwen \VT & 73.1 & 75.8 & 74.5 & 71.5 & 50.6 & 54.3 & 61.8 & 62.3 & 62.1 & 80.7 & 53.1 & 53.7 \\
InternVL \VO & 70.6 & 68.5 & 69.6 & 72.4 & 44.6 & 44.0 & 50.0 & 51.3 & 50.7 & 38.5 & 54.6 & 53.7 \\
InternVL \TO & 55.9 & 56.7 & 56.3 & 57.3 & 46.6 & 49.4 & 64.2 & 64.7 & 64.4 & 66.7 & 64.4 & 64.2 \\
InternVL \VT & 77.0 & 75.4 & 76.2 & 72.6 & 39.4 & 40.9 & 53.6 & 55.7 & 54.7 & 49.8 & 60.0 & 59.4 \\
\addlinespace[2pt]
Sonnet \VO & 60.6 & 63.8 & 62.2 & 52.9 & 44.8 & 44.9 & 50.8 & 52.6 & 51.7 & 32.2 & 81.6 & 81.1 \\
Sonnet \TO & 58.3 & 56.1 & 57.2 & 85.5 & 49.2 & 48.1 & 69.7 & 70.8 & 70.3 & 86.9 & 53.5 & 54.8 \\
Sonnet \VT & 63.5 & 63.1 & 63.3 & 65.2 & 40.9 & 37.9 & 66.9 & 68.0 & 67.4 & 78.6 & 56.5 & 57.2 \\
Opus \VO & 60.1 & 63.7 & 61.9 & 74.4 & 58.7 & 59.5 & 48.7 & 47.2 & 48.0 & 48.8 & 71.9 & 73.6 \\
Opus \TO & 58.0 & 60.5 & 59.2 & 86.2 & 52.9 & 54.0 & 71.2 & 70.3 & 70.8 & 92.0 & 52.3 & 52.9 \\
Opus \VT & 64.9 & 64.4 & 64.7 & 77.3 & 58.2 & 60.8 & 68.7 & 69.1 & 68.9 & 88.1 & 54.9 & 55.3 \\
\addlinespace[2pt]
Fine-tuned \VO & 99.7 & 99.4 & 99.6 & 99.6 & 50.1 & 49.5 & 66.2 & 67.7 & 66.9 & 75.1 & 55.6 & 54.4 \\
Fine-tuned \VT{} (TaP) & 97.6 & 97.8 & 97.7 & 98.0 & 51.0 & 51.0 & 72.9 & 71.7 & 72.3 & 84.6 & 51.3 & 51.6 \\
\bottomrule
\end{tabular}
\caption{\textbf{Order-swap control on relative questions} (\%). Every pair is evaluated in its original order and with the two clips exchanged (the gold letter flipped accordingly). Accuracy columns are subcategory macro-averages, and \emph{avg.} averages the two orders. \emph{Same clip} is the share of pairs for which the same clip is chosen in both orders. A model that always answers by position scores $0$, whereas one that ignores position scores $100$. \emph{B} is the rate of answering B in each order. The original order of the open-weight and fine-tuned models is re-evaluated in the same software environment as the swapped order and can differ slightly from \cref{tab:main}. The proprietary rows use the runs from the main text.}
\label{tab:swap}
\end{table*}

\begin{table*}[tp]
\centering\small
\setlength{\tabcolsep}{3pt}
\begin{tabular}{@{}lrrrrrr@{}}
\toprule
Comparison & $n$ & $a$ & $b$ & $a-b$ & 95\% CI & McNemar $p$ \\
\midrule
Fine-tuned, \VT{} TaP $-$ \VO & 2{,}637 & 73.1 & 67.4 & +5.7 & [+3.0, +8.4] & $4.7\times10^{-13}$ \\
Fine-tuned, \VT{} TaP $-$ \VT{} tokens & 2{,}637 & 73.1 & 67.3 & +5.7 & [+3.3, +8.2] & $1.3\times10^{-12}$ \\
Zero-shot InternVL, \VT{} $-$ \VO & 2{,}637 & 53.6 & 50.0 & +3.6 & [+1.9, +5.3] & $4.0\times10^{-12}$ \\
Zero-shot Qwen, \VT{} $-$ \VO & 2{,}637 & 62.3 & 49.8 & +12.5 & [+10.0, +15.1] & $8.2\times10^{-41}$ \\
Zero-shot Sonnet, \VT{} $-$ \VO & 2{,}637 & 66.9 & 50.8 & +16.0 & [+13.1, +19.1] & $2.9\times10^{-62}$ \\
Zero-shot Opus, \VT{} $-$ \VO & 2{,}637 & 68.7 & 48.7 & +20.0 & [+16.9, +23.0] & $1.6\times10^{-78}$ \\
Zero-shot mean, \VT{} $-$ \VO & 2{,}637 & 62.9 & 49.8 & +13.0 & [+11.4, +14.7] & -- \\
Zero-shot InternVL, \VO{} Tier~A $-$ Tier~B & 543 / 2{,}637 & 70.6 & 50.0 & +20.6 & [+16.3, +24.8] & -- \\
Zero-shot Qwen, \VO{} Tier~A $-$ Tier~B & 543 / 2{,}637 & 74.3 & 49.8 & +24.5 & [+20.5, +28.5] & -- \\
Zero-shot Sonnet, \VO{} Tier~A $-$ Tier~B & 543 / 2{,}637 & 60.6 & 50.8 & +9.8 & [+4.8, +14.6] & -- \\
Zero-shot Opus, \VO{} Tier~A $-$ Tier~B & 543 / 2{,}637 & 60.1 & 48.7 & +11.4 & [+7.1, +15.6] & -- \\
Zero-shot mean, \VO{} Tier~A $-$ Tier~B & 543 / 2{,}637 & 66.4 & 49.8 & +16.5 & [+13.7, +19.4] & -- \\
Fine-tuned, \VT{} TaP $-$ \VO{} (absolute) & 682 & 66.5 & 63.3 & +3.2 & [-1.2, +7.7] & 0.47 \\
\bottomrule
\end{tabular}
\caption{\textbf{Uncertainty of the main comparisons} (relative Tier~B unless noted; macro accuracy, \%). Intervals are percentile intervals from 10{,}000 bootstrap resamples of items, recomputing the subcategory macro in each resample. Paired comparisons resample the shared items jointly, and the four-model mean resamples items jointly across models. McNemar tests are exact and two-sided on the paired item outcomes. They are not defined for the four-model mean or for the Tier~A versus Tier~B comparison, whose items differ (the two tiers are resampled independently).}
\label{tab:stats}
\end{table*}

\begin{table*}[tp]
\centering\small
\setlength{\tabcolsep}{3pt}
\begin{tabular}{@{}lrrrrrrrrrr@{}}
\toprule
& \multicolumn{3}{c}{Relative, Tier~A} & \multicolumn{3}{c}{Relative, Tier~B} & \multicolumn{3}{c}{Absolute, Tier~B} & \\
\cmidrule(lr){2-4}\cmidrule(lr){5-7}\cmidrule(lr){8-10}
Model / input & real & shuf. & withh. & real & shuf. & withh. & real & shuf. & withh. & Shuffle drop (rel.~B) \\
\midrule
Fine-tuned \TO{} (TaP) & 54.3 & 46.6 & 50.1 & 71.7 & 54.5 & 48.4 & 51.8 & 49.1 & 48.4 & +17.2 \\
Fine-tuned \VT{} (TaP) & 97.6 & 90.5 & 98.2 & 72.9 & 59.8 & 63.9 & 65.7 & 65.4 & 66.5 & +13.0 \\
Zero-shot Qwen \TO & 58.2 & 43.5 & 50.1 & 69.8 & 50.1 & 50.8 & 48.9 & 48.2 & 48.4 & +19.7 \\
Zero-shot Qwen \VT & 73.1 & 67.3 & 74.2 & 61.8 & 50.5 & 49.9 & 50.4 & 49.4 & 50.7 & +11.3 \\
Zero-shot InternVL \TO & 55.9 & 42.2 & 48.5 & 64.2 & 49.1 & 49.4 & 48.6 & 49.2 & 48.4 & +15.0 \\
Zero-shot InternVL \VT & 77.0 & 75.8 & 70.6 & 53.6 & 51.3 & 50.0 & 52.3 & 52.0 & 50.2 & +2.3 \\
\bottomrule
\end{tabular}
\caption{\textbf{Tactile controls at inference} (macro accuracy, \%). \emph{Shuffled}: each clip's tactile line is replaced by that of a randomly chosen evaluation clip of the same fold (one fixed assignment, seed 0). \emph{Withheld}: the tactile line is deleted and the rest of the prompt is unchanged. All three conditions were evaluated in a single software environment. This re-evaluation agrees with the predictions behind \cref{tab:main} on 99\% of items for the \VT{} models, 90\% (relative) and 93\% (absolute) for the fine-tuned \TO{} model and 97 to 100\% for the zero-shot models, so the \emph{real} columns differ from \cref{tab:main} by up to $2.5$ points. The shuffle drop is real minus shuffled relative Tier~B accuracy, in points.}
\label{tab:app-controls}
\end{table*}

\begin{table*}[tp]
\centering\small
\setlength{\tabcolsep}{3pt}
\begin{tabular}{@{}lrrrrrrrrrr@{}}
\toprule
& \multicolumn{5}{c}{Relative, Tier~B} & \multicolumn{5}{c}{Absolute, Tier~B} \\
\cmidrule(lr){2-6}\cmidrule(lr){7-11}
Fine-tuned arm & P1 & P2 & P3 & P4 & P5 & P1 & P2 & P3 & P4 & P5 \\
\midrule
\VO & 72.6 & 74.0 & 65.7 & 66.0 & 66.2 & 68.3 & 65.9 & 55.1 & 62.2 & 62.7 \\
\TO{} (TaP) & 65.7 & 78.6 & 47.6 & 67.9 & 73.7 & 52.3 & 48.4 & 47.6 & 52.7 & 48.6 \\
\VT{} (TaP) & 81.5 & 86.0 & 66.0 & 68.9 & 66.2 & 74.4 & 67.3 & 64.1 & 61.2 & 65.5 \\
\VT{} (tokens) & 70.9 & 74.5 & 60.1 & 73.4 & 65.4 & 68.8 & 54.7 & 55.5 & 62.9 & 61.9 \\
\midrule
$n$ (Tier~B) & 344 & 693 & 496 & 371 & 733 & 108 & 166 & 136 & 114 & 158 \\
\bottomrule
\end{tabular}
\caption{\textbf{Per-fold Tier~B macro accuracy} (\%) of the fine-tuned Qwen3.5-9B arms, by held-out participant. Each fold's macro averages the sixteen Tier~B capture subcategories.}
\label{tab:folds}
\end{table*}

\subsection{Per-dimension results and answer priors}
\cref{tab:dim-A,tab:dim-B} report the relative accuracy per scoring dimension for
every model. \Cref{fig:dims} plots the Tier~B zero-shot mean and the
fine-tuned columns. \cref{tab:bias} lists how often each video-only model
answers B, the second clip, against the gold rate. All zero-shot responses
were valid, with none unparsed or missing.

\begin{table*}[tp]
\centering\small
\setlength{\tabcolsep}{2.6pt}
\begin{tabular}{@{}lrrrrrrrrrrrrrrrr@{}}
\toprule
& & \multicolumn{3}{c}{InternVL3-8B} & \multicolumn{3}{c}{Qwen3.5-9B} & \multicolumn{3}{c}{Claude Sonnet} & \multicolumn{3}{c}{Claude Opus} & \multicolumn{3}{c}{Fine-tuned} \\
\cmidrule(lr){3-5}\cmidrule(lr){6-8}\cmidrule(lr){9-11}\cmidrule(lr){12-14}\cmidrule(lr){15-17}
Dimension & $n$ & \VO & \TO & \VT & \VO & \TO & \VT & \VO & \TO & \VT & \VO & \TO & \VT & \VO & \TO & \VT \\
\midrule
Bottle fill level & 84 & 83.3 & 100.0 & 90.5 & 66.7 & 100.0 & 100.0 & 59.5 & 100.0 & 100.0 & 65.5 & 100.0 & 100.0 & 98.8 & 94.0 & 100.0 \\
Touch target & 188 & 88.8 & 48.9 & 92.6 & 84.0 & 50.0 & 72.3 & 74.5 & 44.7 & 71.8 & 54.8 & 46.8 & 61.7 & 100.0 & 43.6 & 96.8 \\
Phone vibration & 45 & 75.6 & 48.9 & 80.0 & 100.0 & 48.9 & 93.3 & 44.4 & 53.3 & 48.9 & 100.0 & 48.9 & 71.1 & 100.0 & 51.1 & 100.0 \\
Texture, material & 85 & 95.3 & 28.2 & 97.6 & 100.0 & 35.3 & 98.8 & 85.9 & 42.4 & 47.1 & 76.5 & 35.3 & 49.4 & 98.8 & 62.4 & 100.0 \\
Texture, structure & 141 & 13.5 & 53.9 & 27.0 & 29.8 & 53.9 & 22.7 & 36.2 & 59.6 & 41.1 & 24.1 & 62.4 & 46.1 & 99.3 & 48.9 & 93.6 \\
\bottomrule
\end{tabular}
\caption{\textbf{Tier~A relative accuracy per scoring dimension} (\%) for zero-shot models and the fine-tuned Qwen3.5-9B (LOUO). Accuracy is item-weighted within each dimension, and $n$ is the number of Tier~A pairs.}
\label{tab:dim-A}
\end{table*}

\begin{table*}[tp]
\centering\small
\setlength{\tabcolsep}{2.6pt}
\begin{tabular}{@{}lrrrrrrrrrrrrrrrr@{}}
\toprule
& & \multicolumn{3}{c}{InternVL3-8B} & \multicolumn{3}{c}{Qwen3.5-9B} & \multicolumn{3}{c}{Claude Sonnet} & \multicolumn{3}{c}{Claude Opus} & \multicolumn{3}{c}{Fine-tuned} \\
\cmidrule(lr){3-5}\cmidrule(lr){6-8}\cmidrule(lr){9-11}\cmidrule(lr){12-14}\cmidrule(lr){15-17}
Dimension & $n$ & \VO & \TO & \VT & \VO & \TO & \VT & \VO & \TO & \VT & \VO & \TO & \VT & \VO & \TO & \VT \\
\midrule
Elastic band & 348 & 56.0 & 70.4 & 70.1 & 73.3 & 74.4 & 79.9 & 58.9 & 76.7 & 71.0 & 59.5 & 75.6 & 75.0 & 83.6 & 73.6 & 78.2 \\
Bottle fill level & 372 & 56.2 & 84.9 & 64.0 & 49.2 & 85.2 & 68.5 & 46.2 & 84.1 & 83.1 & 47.6 & 87.9 & 84.7 & 68.3 & 82.8 & 84.4 \\
Cable seating & 446 & 48.4 & 50.0 & 50.2 & 56.3 & 70.0 & 59.6 & 60.3 & 72.6 & 67.5 & 48.9 & 74.0 & 65.5 & 67.3 & 73.3 & 75.6 \\
Cap tightness & 387 & 56.1 & 80.4 & 54.5 & 56.8 & 74.2 & 73.1 & 51.2 & 74.4 & 76.0 & 50.1 & 78.8 & 78.6 & 70.3 & 76.0 & 80.4 \\
Fruit authenticity & 359 & 42.6 & 79.7 & 46.5 & 50.1 & 78.0 & 62.4 & 52.1 & 79.9 & 78.8 & 50.4 & 81.6 & 83.6 & 81.6 & 81.9 & 81.1 \\
Touch target & 37 & 37.8 & 45.9 & 37.8 & 29.7 & 59.5 & 48.6 & 45.9 & 45.9 & 43.2 & 48.6 & 45.9 & 43.2 & 56.8 & 48.6 & 56.8 \\
Phone vibration & 138 & 51.4 & 50.7 & 50.0 & 49.3 & 50.7 & 48.6 & 49.3 & 50.7 & 47.8 & 49.3 & 52.2 & 54.3 & 55.8 & 47.8 & 47.1 \\
Ball inflation & 121 & 67.8 & 81.8 & 79.3 & 47.1 & 86.0 & 87.6 & 44.6 & 81.0 & 81.0 & 54.5 & 77.7 & 66.1 & 56.2 & 80.2 & 86.0 \\
Pillow filling & 59 & 42.4 & 47.5 & 32.2 & 40.7 & 57.6 & 49.2 & 45.8 & 52.5 & 45.8 & 25.4 & 49.2 & 50.8 & 61.0 & 61.0 & 78.0 \\
Texture, material & 24 & 50.0 & 41.7 & 50.0 & 41.7 & 33.3 & 37.5 & 58.3 & 58.3 & 50.0 & 54.2 & 66.7 & 62.5 & 87.5 & 50.0 & 45.8 \\
Texture, structure & 49 & 24.5 & 34.7 & 30.6 & 28.6 & 67.3 & 36.7 & 28.6 & 67.3 & 55.1 & 24.5 & 75.5 & 71.4 & 65.3 & 38.8 & 55.1 \\
Wind strength & 297 & 60.9 & 82.8 & 72.7 & 45.5 & 65.0 & 77.8 & 50.2 & 69.4 & 73.7 & 60.9 & 70.7 & 73.4 & 62.3 & 79.8 & 82.2 \\
\bottomrule
\end{tabular}
\caption{\textbf{Tier~B relative accuracy per scoring dimension} (\%) for zero-shot models and the fine-tuned Qwen3.5-9B (LOUO). Accuracy is item-weighted within each dimension, and $n$ is the number of Tier~B pairs. \Cref{fig:dims} plots the zero-shot mean and the fine-tuned columns.}
\label{tab:dim-B}
\end{table*}

\begin{table}[tbp]
\centering\small
\setlength{\tabcolsep}{2.6pt}
\begin{tabular}{@{}llrrrrrr@{}}
\toprule
& & & & \multicolumn{4}{c}{\VO{} model answers B} \\
\cmidrule(lr){5-8}
Tier & Dimension & $n$ & Gold B & IVL & Qwen & Son. & Opus \\
\midrule
A & All dimensions & 543 & 50.5 & 44.6 & 56.2 & 44.8 & 58.7 \\
A & Bottle fill level & 84 & 40.5 & 23.8 & 73.8 & 81.0 & 72.6 \\
A & Touch target & 188 & 51.1 & 50.5 & 48.9 & 37.2 & 55.9 \\
A & Phone vibration & 45 & 48.9 & 73.3 & 48.9 & 91.1 & 48.9 \\
A & Texture, material & 85 & 52.9 & 52.9 & 52.9 & 43.5 & 64.7 \\
A & Texture, structure & 141 & 54.6 & 34.8 & 59.6 & 19.1 & 53.9 \\
\midrule
B & All dimensions & 2637 & 49.3 & 54.6 & 71.4 & 81.6 & 71.9 \\
B & Elastic band & 348 & 53.4 & 30.2 & 54.3 & 66.4 & 73.3 \\
B & Bottle fill level & 372 & 47.0 & 12.4 & 69.4 & 94.9 & 84.4 \\
B & Cable seating & 446 & 49.3 & 85.2 & 56.7 & 70.6 & 69.5 \\
B & Cap tightness & 387 & 50.6 & 85.8 & 82.9 & 99.0 & 89.1 \\
B & Fruit authenticity & 359 & 51.0 & 43.7 & 78.6 & 74.9 & 71.0 \\
B & Touch target & 37 & 56.8 & 10.8 & 67.6 & 67.6 & 32.4 \\
B & Phone vibration & 138 & 50.7 & 97.8 & 34.8 & 88.4 & 44.9 \\
B & Ball inflation & 121 & 42.1 & 61.2 & 91.7 & 94.2 & 61.2 \\
B & Pillow filling & 59 & 40.7 & 30.5 & 100.0 & 81.4 & 71.2 \\
B & Texture, material & 24 & 50.0 & 0.0 & 91.7 & 33.3 & 45.8 \\
B & Texture, structure & 49 & 53.1 & 38.8 & 75.5 & 18.4 & 63.3 \\
B & Wind strength & 297 & 45.5 & 56.9 & 93.3 & 92.6 & 62.3 \\
\bottomrule
\end{tabular}
\caption{\textbf{Answer-position frequencies} on relative questions (\%). Model columns use the video-only responses (IVL: InternVL3-8B, Son.: Claude Sonnet). Gold frequencies use all items. Pair orientation was randomized with probability 0.5 before the answer letter was assigned. Frequencies are for the original orientation. \cref{tab:swap} evaluates both orientations.}
\label{tab:bias}
\end{table}

\subsection{Unseen objects}
\label{app:unseen}
\begin{figure*}[tp]
  \centering
  \includegraphics[width=\linewidth]{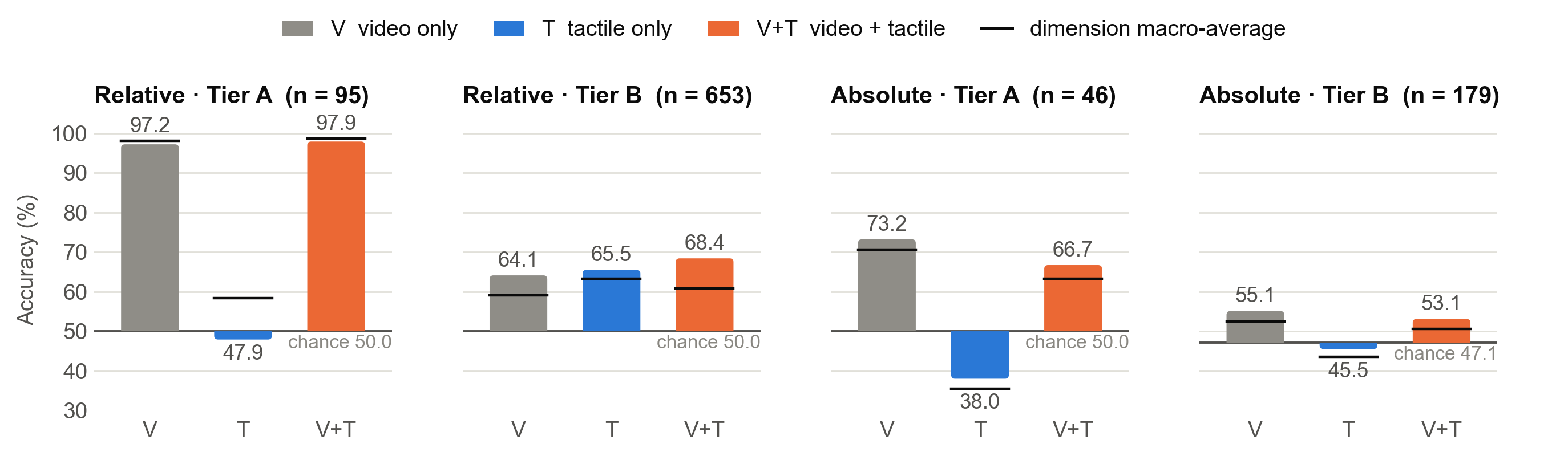}
  \caption{\textbf{Unseen objects with an unseen user, by question format and tier.} Bar height and the printed value show item-weighted
  accuracies averaged over seeds, as in \cref{tab:newobj}. The tick
  marks the unweighted mean over scoring dimensions. On relative comparison,
  adding tactile input helps in both tiers, and on Tier~B tactile-only text
  has the highest dimension macro-average ($63.3$). On absolute questions the
  ordering is reversed. The chance line of the absolute Tier~B panel is the
  item-weighted value $47.1$. Under dimension averaging it is $47.5$.}
  \label{fig:newobj}
\end{figure*}

\Cref{fig:unseen} shows examples of the unseen object instances.
\cref{fig:newobj} plots the unseen-object results by question format and tier,
and \cref{tab:newobj-detail} gives both averages, including the 158
mixed-tier pairs. Under the dimension macro-average, which weights the small
dimensions more heavily, the Tier~B gain of \VT{} over \VO{} is smaller than
under the item-weighted average ($59.1$ to $60.8$ against $64.1$ to $68.4$), and tactile text
alone is highest ($63.3$).

\begin{table*}[tp]
\centering\small
\setlength{\tabcolsep}{3pt}
\begin{tabular}{@{}lrrrrrrrr@{}}
\toprule
Form & Tier & $n$ & Macro V & Macro T & Macro V+T & Micro V & Micro T & Micro V+T \\
\midrule
Relative & A & 95 & 98.1 & 58.3 & 98.6 & 97.2 & 47.9 & 97.9 \\
Relative & B & 653 & 59.1 & 63.3 & 60.8 & 64.1 & 65.5 & 68.4 \\
Relative & mixed & 158 & 80.4 & 59.0 & 87.8 & 78.7 & 60.1 & 88.2 \\
Relative & all & 906 & 71.5 & 61.6 & 75.8 & 70.1 & 62.7 & 74.9 \\
Absolute & A & 46 & 70.6 & 35.5 & 63.2 & 73.2 & 38.0 & 66.7 \\
Absolute & B & 179 & 52.5 & 43.5 & 50.6 & 55.1 & 45.5 & 53.1 \\
Absolute & all & 225 & 57.2 & 41.3 & 54.4 & 58.8 & 44.0 & 55.9 \\
\bottomrule
\end{tabular}
\caption{\textbf{Unseen-object accuracy} (\%). Macro means the unweighted mean across scoring dimensions in each subset. Video-only and video+tactile use three seeds (42--44); tactile-only uses two (42--43). Relative overall includes 158 mixed-tier pairs. Tiers are the corrected assignment (semi-transparent bottle in Tier~B).}
\label{tab:newobj-detail}
\end{table*}

\subsection{Tactile interface}
\begin{figure}[tbp]
  \centering
  \includegraphics[width=\linewidth]{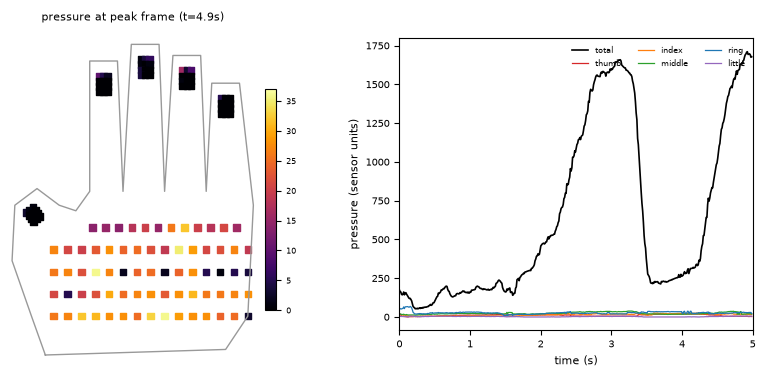}
  \caption{\textbf{Example inputs of the training-free tactile formats} for one ball-squeeze clip. The rendered image shows per-taxel pressure on the glove layout at the frame of peak total force alongside the total and per-finger pressure over the 5\,s clip. The raw-value text for the same clip is: \emph{``readings every 0.2 s over 5 s (sensor units): total pressure 135 59 111 162 157 181 190 187 235 346 500 705 1065 1408 1572 1628 1442 447 225 249 300 525 1030 1477 1665; thumb fingertip 5 4 6 9 11 12 14 14 15 12 14 13 7 5 5 5 6 15 18 19 19 17 11 5 4; index fingertip 6 4 5 6 5 7 6 6 8 12 14 18 18 18 18 18 18 16 23 22 24 21 18 19 17; middle fingertip 15 14 19 21 23 22 22 24 20 19 23 24 31 33 35 34 35 21 18 18 19 22 28 34 21; ring fingertip 58 21 26 27 30 30 26 15 16 18 23 28 25 25 22 20 27 31 34 34 28 28 23 26 29; little fingertip 3 3 4 5 4 4 3 2 1 0 0 1 3 3 4 4 4 0 0 0 1 1 3 4 5''}}
  \label{fig:t16-example}
\end{figure}

\begin{figure}[tbp]
  \centering
  \includegraphics[width=\linewidth]{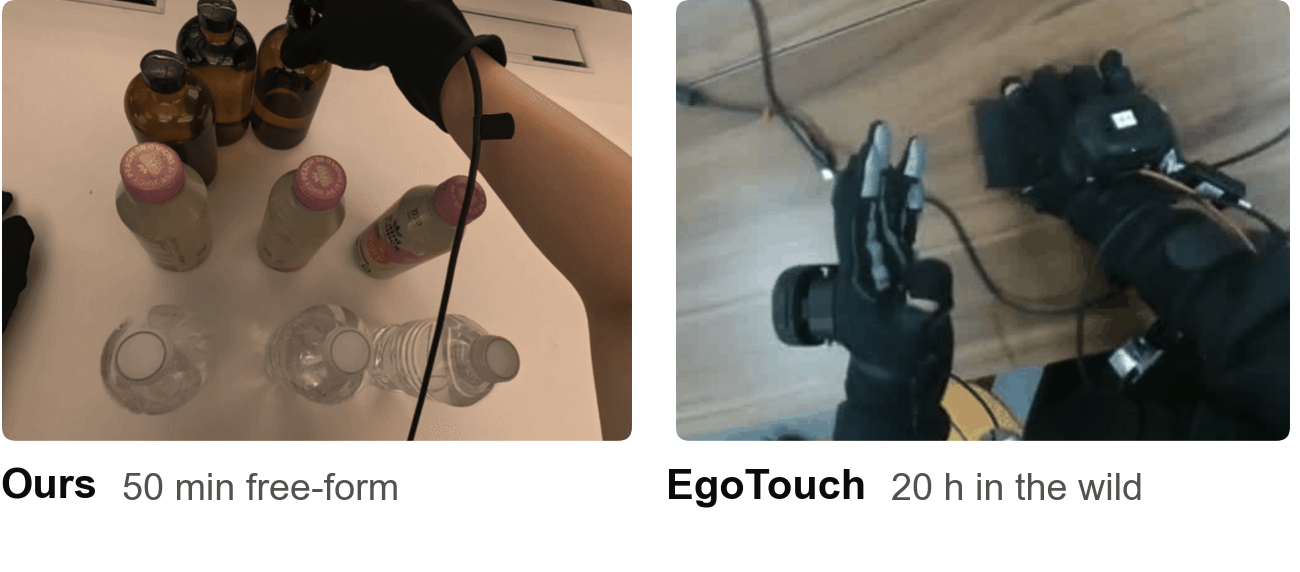}
  \caption{\textbf{Unlabeled data used to pre-train the learned-token tactile
  encoder.} 50 minutes of our own freeform manipulation (left) plus 20 hours
  from a public egocentric tactile dataset (right).}
  \label{fig:pretrain}

\end{figure}

\cref{tab:t16} compares the 34-statistic prompt with two other training-free
formats: a rendered pressure image and raw per-finger pressure as text
(\cref{fig:t16-example}). \cref{tab:app-ablation,tab:ablation-dim} give the
descriptor-group ablation summarized in \cref{sec:ablation}. Removing the spectral
group does not lower phone-vibration accuracy ($77.3\%$ to $85.0\%$ over all
tiers, including the visible face-screen subcategory), consistent with the
glove's 49\,Hz Nyquist limit. \cref{fig:pretrain} shows the unlabeled data used
to pre-train the learned-token encoder.

\begin{table*}[tp]
\centering\small
\setlength{\tabcolsep}{3pt}
\begin{tabular}{@{}llrrrrrr@{}}
\toprule
& & \multicolumn{2}{c}{TaP (34 statistics)} & \multicolumn{2}{c}{Tactile as image} & \multicolumn{2}{c}{Raw values as text} \\
\cmidrule(lr){3-4}\cmidrule(lr){5-6}\cmidrule(lr){7-8}
Model & Input & Tier~A & Tier~B & Tier~A & Tier~B & Tier~A & Tier~B \\
\midrule
Qwen3.5-9B & \TO & 57.3 & 69.7 & 51.2 & 56.4 & 55.0 & 65.7 \\
Qwen3.5-9B & \VT & 73.4 & 62.3 & 70.8 & 51.1 & 75.8 & 58.3 \\
InternVL3-8B & \TO & 55.9 & 64.2 & 55.7 & 57.0 & 56.7 & 60.9 \\
InternVL3-8B & \VT & 77.0 & 53.6 & 75.5 & 53.0 & 74.0 & 51.4 \\
Claude Sonnet & \TO & 58.3 & 69.7 & 61.2 & 70.1 & 54.3 & 62.0 \\
Claude Sonnet & \VT & 63.5 & 66.9 & 66.4 & 64.2 & 63.1 & 61.6 \\
Claude Opus & \TO & 58.0 & 71.2 & 67.9 & 70.9 & 61.0 & 69.4 \\
Claude Opus & \VT & 64.9 & 68.7 & 70.9 & 64.0 & 66.3 & 63.6 \\
\bottomrule
\end{tabular}
\caption{\textbf{Training-free tactile formats}: zero-shot relative accuracy (subcategory macro, \%; 543 Tier~A and 2{,}637 Tier~B pairs). \emph{Tactile as image} replaces each clip's tactile line with one rendered image (\cref{fig:t16-example}). \emph{Raw values as text} replaces the 34 statistics with total and per-finger fingertip pressure averaged over 0.2\,s bins and rounded to integers. Instruction, question and options are unchanged. The TaP columns repeat \cref{tab:main}.}
\label{tab:t16}
\end{table*}

\begin{table}[tbp]
\centering\small
\setlength{\tabcolsep}{3pt}
\begin{tabular}{@{}lrrrrrr@{}}
\toprule
& \multicolumn{4}{c}{Accuracy} & \multicolumn{2}{c}{Shuffle drop} \\
\cmidrule(lr){2-5}\cmidrule(lr){6-7}
Removed & Tier~A & Tier~B & All & B shuf. & B & All \\
\midrule
Full & 97.6 & 72.9 & 83.7 & 59.8 & 13.0 & 14.8 \\
$-$Force & 99.9 & 68.4 & 80.0 & 66.9 & 1.5 & 2.0 \\
$-$Spatial & 98.5 & 72.9 & 83.7 & 60.2 & 12.7 & 14.8 \\
$-$Spectral & 100.0 & 73.3 & 83.8 & 61.4 & 11.9 & 15.3 \\
$-$Wrist & 99.6 & 71.8 & 83.3 & 60.4 & 11.4 & 13.7 \\
\bottomrule
\end{tabular}
\caption{\textbf{Descriptor-group ablation} (fine-tuned Qwen3.5-9B \VT{} TaP,
LOUO over all five folds, retrained once per removed group with the main
recipe; relative accuracy, \%). Tier~A and Tier~B are subcategory macros over
543 and 2{,}637 pairs. \emph{All} is item-weighted over 4{,}102 pairs.
\emph{B shuf.}: Tier~B accuracy when each clip's tactile line is replaced with
that of a randomly chosen evaluation clip from the same fold (one fixed
assignment). \emph{Shuffle drop}: real minus shuffled accuracy, in points. The
\emph{Full} row re-evaluates the main adapters in the software environment of
the ablation runs.}
\label{tab:app-ablation}
\end{table}
\begin{table*}[tp]
\centering\small
\setlength{\tabcolsep}{6pt}
\begin{tabular}{@{}lrrrrrr@{}}
\toprule
Dimension & $n$ & Full & $-$Force & $-$Spatial & $-$Spectral & $-$Wrist \\
\midrule
Elastic band & 348 & 78.2 & 82.5 & 83.6 & 79.0 & 81.6 \\
Bottle fill level & 804 & 90.2 & 79.5 & 89.2 & 87.8 & 87.3 \\
Cable seating & 446 & 75.6 & 70.4 & 73.8 & 73.3 & 73.3 \\
Cap tightness & 387 & 81.1 & 73.9 & 82.4 & 87.3 & 84.8 \\
Fruit authenticity & 359 & 81.6 & 81.1 & 83.6 & 80.5 & 80.5 \\
Touch target & 389 & 91.5 & 91.5 & 88.4 & 92.5 & 90.7 \\
Phone vibration & 339 & 77.3 & 76.4 & 75.2 & 85.0 & 78.5 \\
Ball inflation & 121 & 86.0 & 75.2 & 88.4 & 77.7 & 76.9 \\
Pillow filling & 59 & 78.0 & 69.5 & 84.7 & 69.5 & 79.7 \\
Texture, material & 199 & 93.5 & 93.5 & 96.0 & 95.5 & 93.0 \\
Texture, structure & 354 & 83.3 & 93.2 & 81.4 & 85.9 & 88.4 \\
Wind strength & 297 & 81.8 & 67.7 & 81.8 & 75.8 & 77.4 \\
\bottomrule
\end{tabular}
\caption{\textbf{Descriptor-group ablation per scoring dimension} (relative accuracy, \%, all tiers including mixed-tier pairs, item-weighted), for the same retrained models as \cref{tab:app-ablation}.}
\label{tab:ablation-dim}
\end{table*}

\subsection{Describing what the camera cannot see}
\label{app:caption}
Multiple-choice evaluation measures whether a model can select the correct answer once
the target property is named in the prompt. We also ask whether the model spontaneously mentions a property that the camera cannot see. A strong proprietary model conditioned on our hand-annotated object properties produces the reference captions, and Qwen3.5-9B describes each clip in its own words.

 In the zero-shot setting, tactile input helps on all three
metrics (BLEU-4 $+0.0083$, ROUGE-L $+0.0180$, CIDEr-D $+0.0339$), and the gain occurs in both tiers (\cref{tab:caption}). After fine-tuning on 650--726 examples per fold, the effect largely disappears ($+0.0010$, $-0.0032$, $-0.0198$). This result is consistent with limited training data, but it does not
establish whether template reproduction is the mechanism. We treat
CIDEr-D as indicative only, since its seed spread of $0.047$ exceeds every
effect reported here.

The 43 excluded clips all belong to the cap-tightness class ``Just screwed
on'', for which no reference caption exists. Held-out caption counts for P1--P5 are 146, 207, 172, 131 and 201, leaving 711, 650, 685, 726 and 656 training examples, respectively.
Zero-shot tactile shuffling changes BLEU-4 far less than withholding the
tactile block, so the modality gain should not be interpreted as proof of
content grounding. After fine-tuning, shuffling the tactile input even raises the multimodal
BLEU-4 slightly ($+0.0007$), so the fine-tuned scores do not show tactile
grounding either. Because the references are written by a proprietary model,
this evaluation also partly measures agreement with that model's phrasing.

\paragraph{A control for training-side changes}
A span-weighted caption loss raised BLEU-4 for the video-only arm from $.0747$
to $.0791$ and for the tactile arm from $.0757$ to $.0787$. It is therefore a
general training improvement rather than a tactile one. Had the improved
tactile arm been scored against the original video-only baseline, it would
have shown $+0.0040$ and been recorded as a gain from touch. The same caution
applies to any modality-injection result whose training recipe changes
together with the modality.

\begin{table*}[tp]
\centering\small
\setlength{\tabcolsep}{4pt}
\begin{tabular}{@{}ll rrr rrr rrr@{}}
\toprule
& & \multicolumn{3}{c}{BLEU-4} & \multicolumn{3}{c}{ROUGE-L} & \multicolumn{3}{c}{CIDEr-D} \\
\cmidrule(lr){3-5}\cmidrule(lr){6-8}\cmidrule(lr){9-11}
Arm & Tier & \VO & \VT & $\Delta$ & \VO & \VT & $\Delta$ & \VO & \VT & $\Delta$ \\
\midrule
\multirow{3}{*}{Zero-shot} & Tier~A & 1.59 & 2.90 & +1.31 & 19.88 & 21.39 & +1.51 & 7.82 & 11.98 & +4.16 \\
 & Tier~B & 1.63 & 2.32 & +0.69 & 18.76 & 20.65 & +1.89 & 9.24 & 12.37 & +3.13 \\
 & all & 1.63 & 2.47 & +0.83 & 19.04 & 20.84 & +1.80 & 9.28 & 12.68 & +3.39 \\
\midrule
\multirow{3}{*}{Fine-tuned} & Tier~A & 7.75 & 7.72 & $-$0.03 & 33.12 & 32.68 & $-$0.44 & 43.06 & 40.86 & $-$2.20 \\
 & Tier~B & 7.38 & 7.52 & +0.14 & 32.82 & 32.54 & $-$0.28 & 46.36 & 44.64 & $-$1.71 \\
 & all & 7.47 & 7.57 & +0.10 & 32.90 & 32.57 & $-$0.32 & 47.37 & 45.39 & $-$1.98 \\
\bottomrule
\end{tabular}
\caption{\textbf{Caption scores} ($\times100$), pooled across held-out folds. Differences are absolute score differences on the same reference set. Tier~A has 218 clips, Tier~B 639 and all 857 (corrected tiers). BLEU-4 uses pooled n-gram counts and ROUGE-L is averaged over items. CIDEr-D recomputes document frequencies within each reported reference set, so tier scores need not average to the all score.}
\label{tab:caption}
\end{table*}

\begin{figure}[tbp]
\centering
\includegraphics[width=\linewidth]{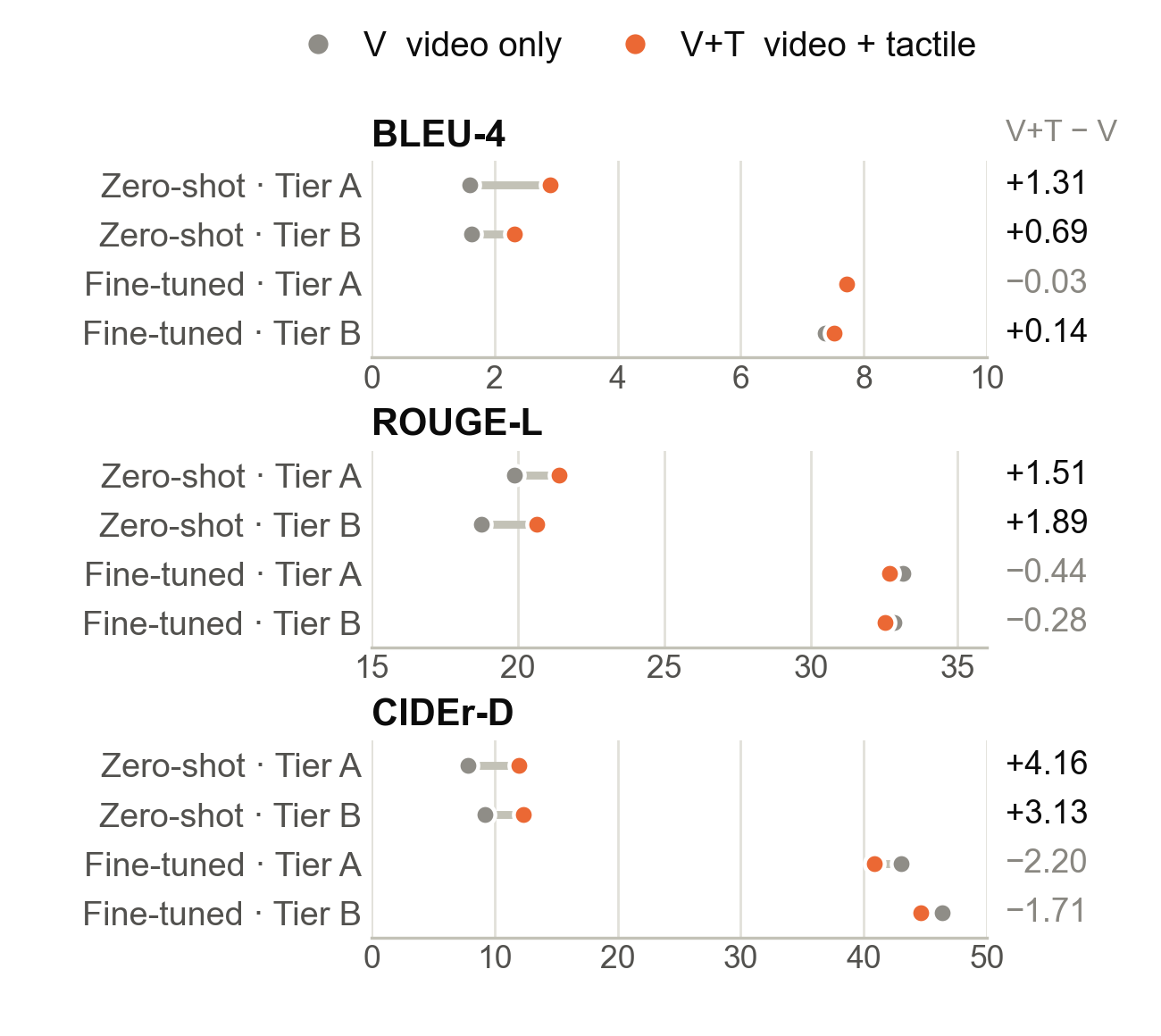}
\caption{Caption metrics for the zero-shot and fine-tuned arms, by tier.
Scores are multiplied by 100 and computed from the same saved predictions
as \cref{tab:caption}.}
\label{fig:caption}
\end{figure}

\FloatBarrier
\section{Web-scale contact QA}
\label{app:web}

\begin{figure*}[tp]
  \centering
  \includegraphics[width=0.92\linewidth]{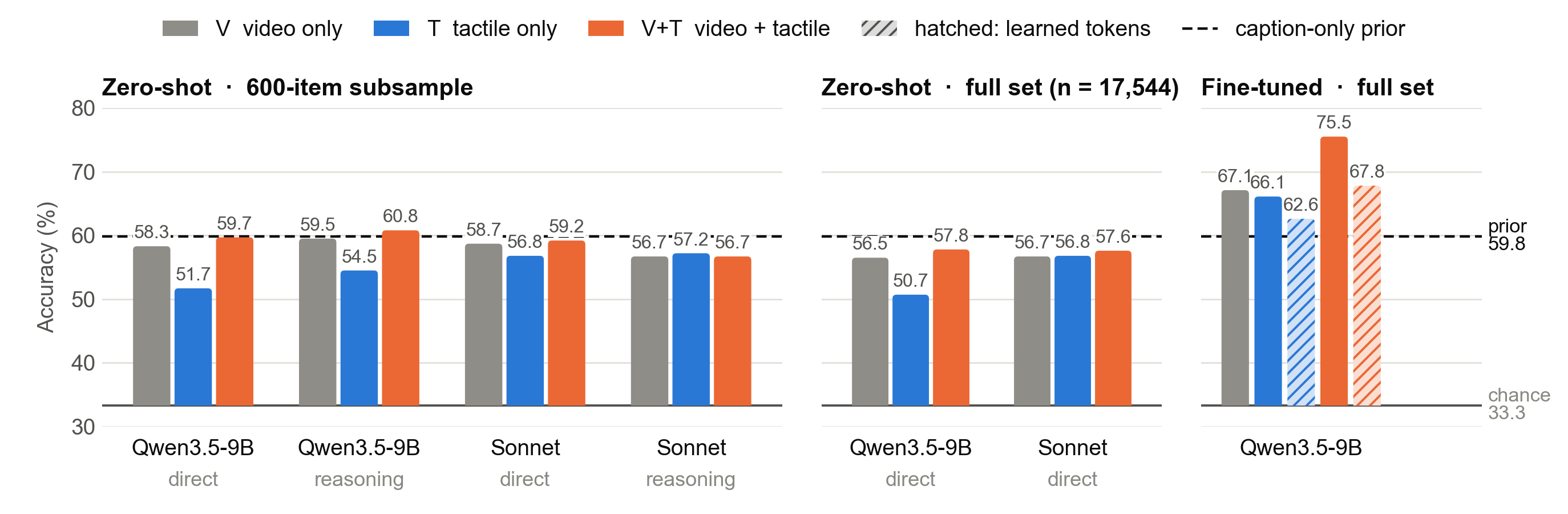}
  \caption{\textbf{Annotation-free contact QA over 20\,h of public egocentric
  video.} The strongest reported zero-shot arm is Qwen3.5-9B with
  reasoning-style prompting and tactile input, scoring $60.8\%$ against the $59.8\%$
  caption-wording prior, a $1.0$-point difference
  on the 600-item subsample. Fine-tuning with tactile input serialized as
  text clears the prior by a meaningful margin, reaching $75.5\%$. The same
  tactile clips injected as learned tokens reach $67.8\%$, barely above the
  $67.1\%$ video-only baseline. Chance is $33.3\%$. ``Direct'' and
  ``reasoning'' denote the two zero-shot prompting styles. Reasoning-style
  prompting was run only on the 600-item subsample, which is stratified to
  120 evaluation items per property and consists of an earlier 120-item
  subset extended by 480 items.}
  \label{fig:egotouch}
\end{figure*}

\cref{fig:egotouch} gives the full web-scale results, including the zero-shot
reasoning-style runs on the 600-item subsample, and \cref{tab:web-clean}
re-scores every arm on subsets that remove the construction shortcuts described
below.

\begin{table*}[tp]
\centering\small
\setlength{\tabcolsep}{3pt}
\begin{tabular}{@{}lrrrrr@{}}
\toprule
Arm & Full & Same hand & No overlap & No ``both'' & All three \\
\midrule
$n$ & 17{,}544 & 5{,}687 & 13{,}242 & 12{,}349 & 4{,}193 \\
\midrule
Fine-tuned \VO & 67.1 & 63.2 & 65.8 & 66.0 & 63.2 \\
Fine-tuned \TO{} (text) & 66.1 & 63.2 & 64.7 & 65.6 & 62.7 \\
Fine-tuned \VT{} (text) & 75.5 & 72.3 & 74.1 & 75.0 & 72.0 \\
Fine-tuned \TO{} (tokens) & 62.6 & 59.4 & 61.2 & 62.0 & 59.8 \\
Fine-tuned \VT{} (tokens) & 67.8 & 63.8 & 66.6 & 66.9 & 64.0 \\
Zero-shot Qwen \VO & 56.5 & 55.5 & 56.3 & 56.2 & 55.9 \\
Zero-shot Qwen \VT{} (text) & 57.8 & 57.4 & 57.9 & 57.4 & 57.5 \\
Zero-shot Qwen \TO{} (text) & 50.7 & 48.6 & 49.8 & 50.0 & 48.4 \\
Zero-shot Sonnet \VO & 56.7 & 54.7 & 56.2 & 56.7 & 55.4 \\
Zero-shot Sonnet \VT{} (text) & 57.6 & 56.0 & 57.0 & 57.5 & 56.7 \\
Zero-shot Sonnet \TO{} (text) & 56.8 & 53.6 & 54.8 & 57.2 & 54.4 \\
Caption-only prior$^\ast$ & 59.8 & 57.1 & 59.8 & 60.0 & 57.4 \\
\bottomrule
\end{tabular}
\caption{\textbf{Web-scale accuracy} (\%) on the full evaluation set and on subsets that remove the construction shortcuts: pairs of events performed with the same hand, pairs whose frame intervals do not overlap, pairs with no ``both'' in either caption and pairs satisfying all three. \emph{Text} is the per-frame tactile table of \cref{sec:scale}, not the 34 statistics. Zero-shot rows use direct prompting, and the Sonnet \VT{} and \TO{} rows exclude 8 and 9 missing responses, respectively. $^\ast$The caption-only prior is retrained on the training items that satisfy the same filter. Of the 11{,}857 mixed-handedness evaluation items, 6{,}227 pair a one-handed with a two-handed event, and the two-handed event is the correct answer in 3{,}638 of them ($30.7\%$ of all mixed-handedness items).}
\label{tab:web-clean}
\end{table*}

\paragraph{Question construction}
A proprietary model of the Claude Opus family segments each recording into
events and captions each event. The caption files record the model family but
not a dated snapshot. A question compares two captioned events
within a five-second clip on one of five contact properties computed from the
raw tactile signal (mean force, peak force, mean contact area, peak contact
area or contact duration). ``About the same'' is the correct third option
when the two values differ by less than $8\%$ of the larger one, or, for
duration, by at most one of 32 frames. The split is by recording, with
1{,}149 training and 206 evaluation recordings. The direct prompts were fixed
in advance, but the answer-extraction rule and generation limits of the
reasoning-style runs, as well as the reasoning prompt of the proprietary model, were
set after inspecting the results on the first 120 items of the 600-item subsample. On the 480 items added afterward, the strongest arm scores
$60.2\%$. An example asks, ``Which action was performed with the greater average
grip force? If an action uses both hands, average the two hands.'' Its options
are ``A) grasp the edge of the grey foam floor mat and rotate it on the blue
table'', ``B) About the same'' and ``C) rest the hand on the rim of the blue
table''. The measured answer is A. The recorded actions use different hands
and overlap in time, illustrating the construction limitations below.

\paragraph{The handedness shortcut}
An earlier construction of the web-scale split paired clips without
constraining handedness. Because that construction aggregated force and contact area
over both hands, two-handed events almost always exceeded
one-handed ones. Mixed-handedness pairs made up $35.6\%$ of the evaluation
set, and the two-handed clip was the correct answer $81.9\%$ of the time, so always
choosing the two-handed option would have scored $82\%$ on a third of the
benchmark. Moreover, $49\%$ of the corresponding captions contained the word
``both''. Accuracy split by pair type confirmed the effect, with $83.6\%$ on
mixed pairs against $68.7\%$ and $62.3\%$ on same-handedness pairs. These
percentages describe the earlier construction and should not be attributed
to the current evaluation set. The current question wording specifies
averaging over the two hands (see the example above), but the construction
does not filter on handedness, temporal overlap or wording. Among the 104{,}210 training items, 68{,}452 mix
handedness, 24{,}268 have overlapping intervals and 27{,}335 mention
``both''. Among the 17{,}544 evaluation items, the corresponding counts are
11{,}857, 4{,}302 and 5{,}195. The recording-level split prevents
cross-recording leakage but does not remove these within-item shortcuts.

\balance %
\paragraph{Tactile controls and the caption prior}
On a 4,000-item evaluation subset, the fine-tuned tactile-text model scores
$75.6\%$ with its own tactile text, $49.7\%$ when each clip's tactile text is
replaced by that of another clip, and $58.9\%$ when it is withheld. The
learned-token model scores $67.4\%$ in all three conditions, indicating that it does not
use its tactile input. Its encoder was trained from scratch on the raw taxels
of both hands, without a gate and without the pre-training used on
TouchBench. Over 200 evaluation clips its tactile tokens have a mean L2 norm
of $57.5$, about 67 times the mean norm of the rows of the language model's
input-embedding matrix ($0.85$), a likely reason the model ignores them. On this corpus the tactile text is a 32-frame
table of left- and right-hand grip force, contact area and contact flag. The
caption-only prior is a three-class softmax on bag-of-words differences
between the two captions, trained separately for each property on the
104,210 training questions (minimum token count 20, three epochs, learning
rate 0.05, seed 0). It answers $10{,}491/17{,}544$ ($59.8\%$) evaluation
items and $359/600$ ($59.8\%$) subsample items correctly.

\FloatBarrier
\clearpage\onecolumn   %
\section{Prompt templates}
\label{app:prompts}

In the listings below, \texttt{\{...\}} marks a runtime insertion and
\texttt{[32 video frames]} marks where the frame tokens are spliced in. A
physical line ending in a backslash is a presentational wrap only: rejoin it
with the following line using a single space, since every logical line reaches
the model unbroken. The question and option text is identical across
conditions, while the system instructions describe the supplied modalities.
The frames are dropped for tactile-only input, and the \texttt{Tactile:} line
appears only when tactile input is provided. Proprietary runs receive the
32 frames of each clip as two contact-sheet images in the message, with no tools.

One wording note: the tactile-only system line says ``tactile sensor tokens''
for historical reasons, but in the Tactile-as-Prompt condition the model
receives the serialized statistics as plain text, exactly as shown below. We
reproduce the prompts verbatim as they were used at evaluation time rather than
tidying them retrospectively.

\subsection{Absolute task: one clip, direct property question}

\paragraph{Vision only (VO)}
\begin{verbatim}
You are a tactile-visual AI analyzing egocentric video of a\
right hand with a tactile glove. You see 32 first-person\
frames.

[32 video frames]
{question}
{option A}
{option B}
{option C}
{option D}

Answer with ONLY the letter (A, B, C, or D).
\end{verbatim}

\paragraph{Tactile only (TO)}
\begin{verbatim}
You are a tactile AI analyzing a right hand with a tactile\
glove. You see tactile sensor tokens encoding fingertip+palm\
pressure and per-finger bend over time.

Tactile: {34-statistic block, serialized as below}
{question}
{option A}
{option B}
{option C}
{option D}

Answer with ONLY the letter (A, B, C, or D).
\end{verbatim}

\paragraph{Vision and tactile (VT)}
\begin{verbatim}
You are a tactile-visual AI analyzing egocentric video of a\
right hand with a tactile glove. You see 32 first-person\
frames, then a summary of the tactile-glove sensor readings.

[32 video frames]
Tactile: {34-statistic block, serialized as below}
{question}
{option A}
{option B}
{option C}
{option D}

Answer with ONLY the letter (A, B, C, or D).
\end{verbatim}

\subsection{Relative task: two clips, comparison}

The preamble is identical across conditions; VO omits the two
\texttt{Tactile:} lines and TO omits the two \texttt{[32 video frames]}
blocks. The variant shown is vision and tactile (VT).
\begin{verbatim}
You are comparing TWO clips of a gloved right hand. CLIP 1\
is shown first, then CLIP 2. Answer the multiple-choice\
question.

CLIP 1:
[32 video frames]
Tactile: {34-statistic block for clip 1}
CLIP 2:
[32 video frames]
Tactile: {34-statistic block for clip 2}
{question}
{option A}
{option B}

Answer with ONLY the letter A or B.
\end{verbatim}

\subsection{Tactile block format}

The 34 statistics are serialized on a single line, prefixed by
\texttt{Tactile:}, in the group order force, spatial, spectral, wrist, as
clauses joined by ``\texttt{;\ }''. Each value below is a placeholder
\texttt{\{name:fmt\}} whose \texttt{fmt} is the exact rounding applied;
quantities carry the units of Table~\ref{tab:features} and appear in the fixed
order shown.
\begin{verbatim}
force peak {f_peak:.0f} / p95 {f_p95:.0f} / mean\
{f_mean:.0f} / sd {f_std:.0f}, strongest single taxel\
{max_single:.0f}; per-finger peak\
{finger0:.0f}/{finger1:.0f}/{finger2:.0f}/{finger3:.0f}/{finger4:.0f},\
fingertip total {fingertip_total:.0f}, palm {palm_force:.0f},\
fingertip-to-palm ratio {fp_ratio:.2f}, finger bend\
{flex_sum:.1f}; contact area {active_count:.0f} taxels,\
concentration gini {gini:.2f} / top1 {top1_frac:.2f} / top3\
{top3_frac:.2f}, spatial cv {spatial_cv:.2f}; {n_peaks:.0f}\
force pulses, fastest rise {f_slope:.0f}, peak at\
{peak_time_frac:.2f} of the clip; vibration power 8-16Hz mean\
{spec_b0816_mean:.2f} max {spec_b0816_max:.2f}, 16-32Hz mean\
{spec_b1632_mean:.2f} max {spec_b1632_max:.2f}, 32-49Hz mean\
{spec_b3249_mean:.2f} max {spec_b3249_max:.2f}, 44-49Hz mean\
{spec_b4449_mean:.2f} max {spec_b4449_max:.2f}, spectral\
centroid {spec_centroid:.2f} Hz, high-frequency jitter\
{hf_diff_rms:.3f}; wrist jitter {wrist_jitter:.4f}, wrist tilt\
range {wrist_tilt_range:.2f}
\end{verbatim}

\end{document}